\documentclass[10pt,twocolumn,letterpaper]{article}

\usepackage{cvpr}              

\usepackage{amsmath}
\usepackage{amssymb}
\usepackage{bm}
\usepackage{booktabs}
\usepackage{enumitem}
\usepackage{gensymb}  
\usepackage{graphicx}
\usepackage{mathtools}  
\usepackage{multirow}
\usepackage{mwe}
\usepackage{placeins}
\usepackage{tabularx}
\usepackage{xcolor}
\usepackage{url}

\graphicspath{ {images/} }
\newcommand{\R}{\mathbb{R}}
\newcommand{\E}{\mathbb{E}}
\DeclareMathOperator*{\argmin}{arg\,min}
\DeclarePairedDelimiter\norm{\lVert}{\rVert}
\DeclarePairedDelimiter\floor{\lfloor}{\rfloor}

\usepackage[pagebackref,breaklinks,colorlinks]{hyperref}

\usepackage[capitalize]{cleveref}
\crefname{section}{Sec.}{Secs.}
\Crefname{section}{Section}{Sections}
\Crefname{table}{Table}{Tables}
\crefname{table}{Tab.}{Tabs.}

\def\cvprPaperID{****} 
\def\confName{CVPR}
\def\confYear{2022}

\begin{document}

\title{REGTR: End-to-end Point Cloud Correspondences with Transformers}

\author{Zi Jian Yew \qquad Gim Hee Lee\\
National University of Singapore\\
{\tt\small yewzijian@u.nus.edu \qquad gimhee.lee@comp.nus.edu.sg}
}
\maketitle

\begin{abstract}
Despite recent success in incorporating learning into point cloud registration, many works focus on learning feature descriptors and continue to rely on nearest-neighbor feature matching and outlier filtering through RANSAC to obtain the final set of correspondences for pose estimation.
In this work, we conjecture that attention mechanisms can replace the role of explicit feature matching and RANSAC, and thus propose an end-to-end framework to directly predict the final set of correspondences. We use a network architecture consisting primarily of transformer layers containing self and cross attentions, and train it to predict the probability each point lies in the overlapping region and its corresponding position in the other point cloud.
The required rigid transformation can then be estimated directly from the predicted correspondences without further post-processing.
Despite its simplicity, our approach achieves state-of-the-art performance on 3DMatch and ModelNet benchmarks. Our source code can be found at \url{https://github.com/yewzijian/RegTR}.
\end{abstract}

\section{Introduction}
Rigid point cloud registration refers to the problem of finding the optimal rotation and translation parameters that align two point clouds. A common solution to point cloud registration follows the following pipeline: 1) detect salient keypoints, 2) compute feature descriptors for these keypoints, 3) obtain putative correspondences via nearest neighbor matching, and 4) estimate the rigid transformation, typically in a robust fashion using RANSAC. In recent years, researchers have applied learning to point cloud registration. Many of these works focus on learning the feature descriptors \cite{zeng20163dmatch,deng2018ppfnet,choy2019fcgf} and sometimes also the keypoint detection \cite{yew20183dfeatnet,bai2020d3feat,huang2021predator}. The final two steps generally remain unchanged and these approaches still require nearest neighbor matching and RANSAC to obtain the final transformation. These algorithms do not take the post-processing into account during training, and their performance can be sensitive to the post-processing choices to pick out the correct correspondences, \eg number of sampled interest points or distance threshold in RANSAC. 

\begin{figure}[t]
    \centering
    \includegraphics[width=0.9\linewidth]{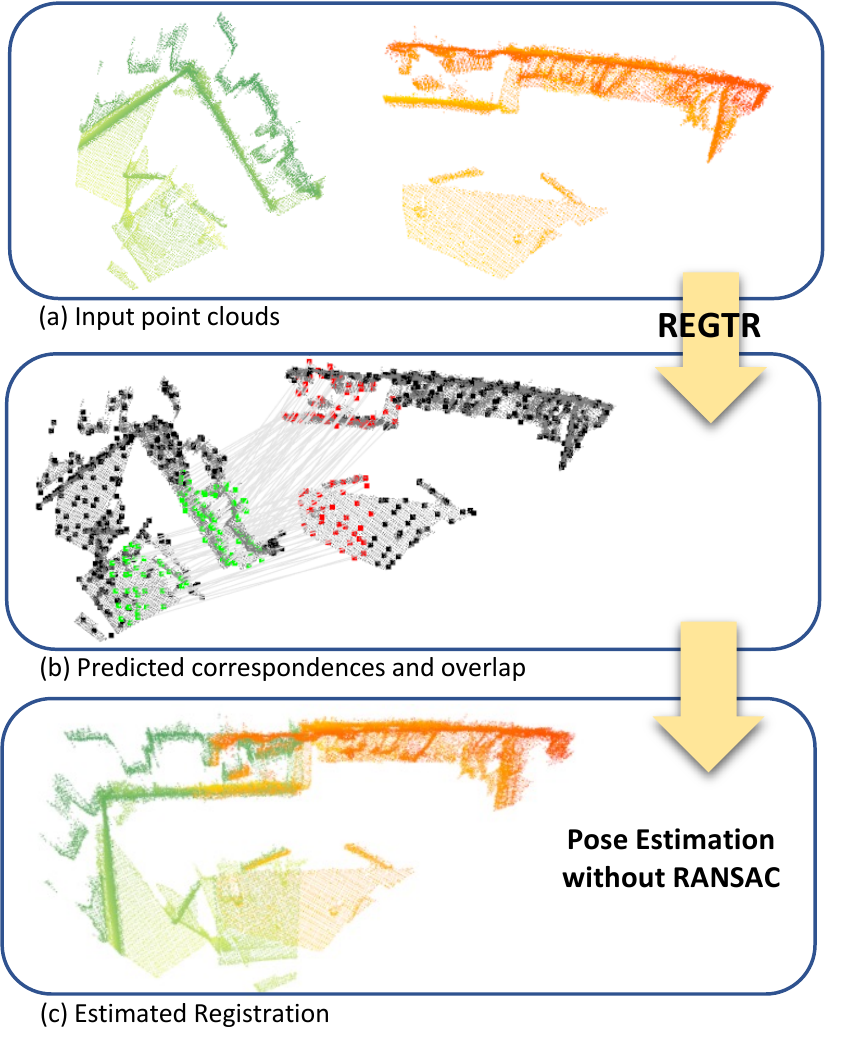}
    \vspace{-3mm}
    \caption{Our network directly outputs final correspondences and the overlap scores. The required rigid transformation can then be directly computed from these correspondences without RANSAC.} \vspace{-5mm}
    \label{fig:teaser}
\end{figure}

Several works \cite{wang2019dcp,yew2020rpmnet,wang2019prnet} avoid the non-differentiable nearest neighbor matching and RANSAC steps by estimating the alignment using soft correspondences computed from the local feature similarity scores. In this work, we take a slightly different approach. 
We observe that the learned local features in these works are mainly used to establish correspondences.
Thus, we focus on having the network directly predict a set of clean correspondences instead of learning good features.
We are motivated by the recent line of works \cite{carion2020detr,misra2021-3detr} which make use of transformer attention \cite{vaswani2017attention} layers to predict the final outputs for various tasks with minimal post-processing. Although attention mechanisms have previously been used in registration of both point clouds \cite{wang2019dcp,huang2021predator} and images \cite{sarlin2020superglue}, these works utilize attention layers mainly to aggregate contextual information to learn more discriminative feature descriptors. A subsequent RANSAC or optimal transport step is still often used to obtain the final correspondences.
In contrast, our Registration Transformer (REGTR) utilizes attention layers to directly output a consistent set of \emph{final} point correspondences, as illustrated in \cref{fig:teaser}. Since our network outputs clean correspondences, the required rigid transformation can be estimated directly without additional nearest neighbor matching and RANSAC steps.

Our REGTR first uses a point convolutional backbone \cite{thomas2019kpconv} to extract a set of features while downsampling the input pair of point clouds. The features of both point clouds are passed into several transformer \cite{vaswani2017attention} layers consisting of multi-head self and cross attentions to allow for global information aggregation, while taking into account the point positions through positional encodings to allow the network to utilize rigidity constraints to correct bad correspondences. 
The resulting features are then used to predict the corresponding transformed locations of the downsampled points.
We additionally predict overlap probability scores to weigh the predicted correspondences when computing the rigid transformation. 
Unlike the more common approach of computing correspondences via nearest neighbor feature matching, which requires interest points to be present at the same locations in both point clouds, our network is trained to directly predict corresponding point locations. As a result, we do not require sampling large number of interest points (\eg in \cite{zeng20163dmatch,huang2021predator}) or a keypoint detector (\eg \cite{zhong2009iss,li2019usip}) that produces repeatable points. Instead, we establish correspondences on simple grid subsampled points.

Although our REGTR is simple in design, it achieves state-of-the-art performance on the 3DMatch \cite{zeng20163dmatch} and ModelNet \cite{wu2015modelnet} datasets. It also has fast run times since it does not require running RANSAC on a large number of putative correspondences. 
In summary, our contributions are:
\vspace{-2mm}
\begin{itemize}
\setlength\itemsep{-0.1em}
    \item We directly predict a consistent set of final point correspondences via self and cross attention, without using the commonly used RANSAC nor optimal transport layers.
    \item We evaluate on several datasets and demonstrate state-of-the-art performance, achieving precise alignments despite using a small number of correspondences.
\end{itemize}
\vspace{-2mm}

\section{Related Work}

\paragraph{Correspondence-based registration.}
Correspondence-based approaches for point cloud registration first establish correspondences between salient keypoints, followed by robust estimation of the rigid transformation. To accomplish the first step, many keypoint detectors \cite{zhong2009iss,steder2010narf} and feature descriptors \cite{rusu2009fpfh,tombari2010usc} have been handcrafted. Pioneered by 3DMatch \cite{zeng20163dmatch}, many researchers propose to improve feature descriptors \cite{zeng20163dmatch,khoury2017cgf,deng2018ppfnet,choy2019fcgf} and also keypoint detection \cite{yew20183dfeatnet,li2019usip,bai2020d3feat} by learning from data. Recently, Predator \cite{huang2021predator} utilizes attention mechanisms to aggregate contextual information to learn more discriminative feature descriptors. Most of these works are trained by optimizing a variant of the contrastive loss \cite{chopra2005contrastiveloss,schroff2015facenet} between feature descriptors of matching and non-matching points, and rely on a subsequent nearest neighbor matching step and RANSAC to select the correct correspondences.

\vspace{-4.2mm}
\paragraph{Learned direct registration methods.}
Instead of combining learned descriptors with robust pose estimation, some works incorporate the entire pose estimation into the training pipeline.
Deep Closest Point (DCP) \cite{wang2019dcp} proposes a learned version of Iterative Closest Point (ICP) \cite{besl1992icp,chen-medioni1991icp}, and utilizes soft correspondences on learned pointwise features to compute the rigid transform in a differentiable manner. However, DCP cannot handle partial overlapping point clouds and thus later works overcome the limitation by detecting keypoints \cite{wang2019prnet} or using optimal transport layers with an added slack row and column \cite{yew2020rpmnet,fischer2021stickypillars}. IDAM \cite{li2020idam} considers both feature and Euclidean space during its pairwise matching process.
PCAM \cite{cao2021pcam} multiplies the cross-attention matrices at multiple levels to fuse low and high-level contextual information.
DeepGMR \cite{yuan2020deepgmr} learns to compute point-to-distribution correspondences.
A separate group of works \cite{aoki2019pointnetlk,li2021pointnetlk2,xu2021omnet,huang2020featuremetric,sarode2019pcrnet} rely on global feature descriptors and circumvent the local feature correspondence step. PointNetLK \cite{aoki2019pointnetlk} aligns two point clouds by minimizing the distances between their global PointNet \cite{qi2017pointnet} features, in a procedure similar to the Lucas-Kanade \cite{baker2004lucas} algorithm.
Li \etal \cite{li2021pointnetlk2} extends it to use analytical Jacobians to improve the generalization behavior. OMNet \cite{xu2021omnet} incorporates masking into the global feature to better handle partial overlapping point clouds.
Our method utilizes local features and is similar to \eg \cite{wang2019dcp,wang2019prnet}, but we focus on predicting accurate corresponding point locations via transformer attention layers.

\vspace{-4.2mm}
\paragraph{Learned correspondence filtering.}
The putative correspondences obtained from correspondence-based methods contain outliers, and thus RANSAC is typically used to filter out wrong matches when estimating the required transformation. However, RANSAC is non-differentiable and cannot be used within a training pipeline. Recent works alleviate this problem by modifying RANSAC to enforce differentiability \cite{brachmann2017dsac}, or by learning to identify which of the putative correspondences are inliers \cite{yi2018learning,choy2020dgr,lee2021houghvote,gojcic2020multiview}. In addition to inlier classification, 3DRegNet \cite{pais20203dregnet} also regresses the rigid transformation parameters using a deep network.
Different from these works, we directly predict the clean correspondences without explicitly computing the noisy putative correspondences. 

\begin{figure*}[t]
    \centering
    \includegraphics[width=\linewidth]{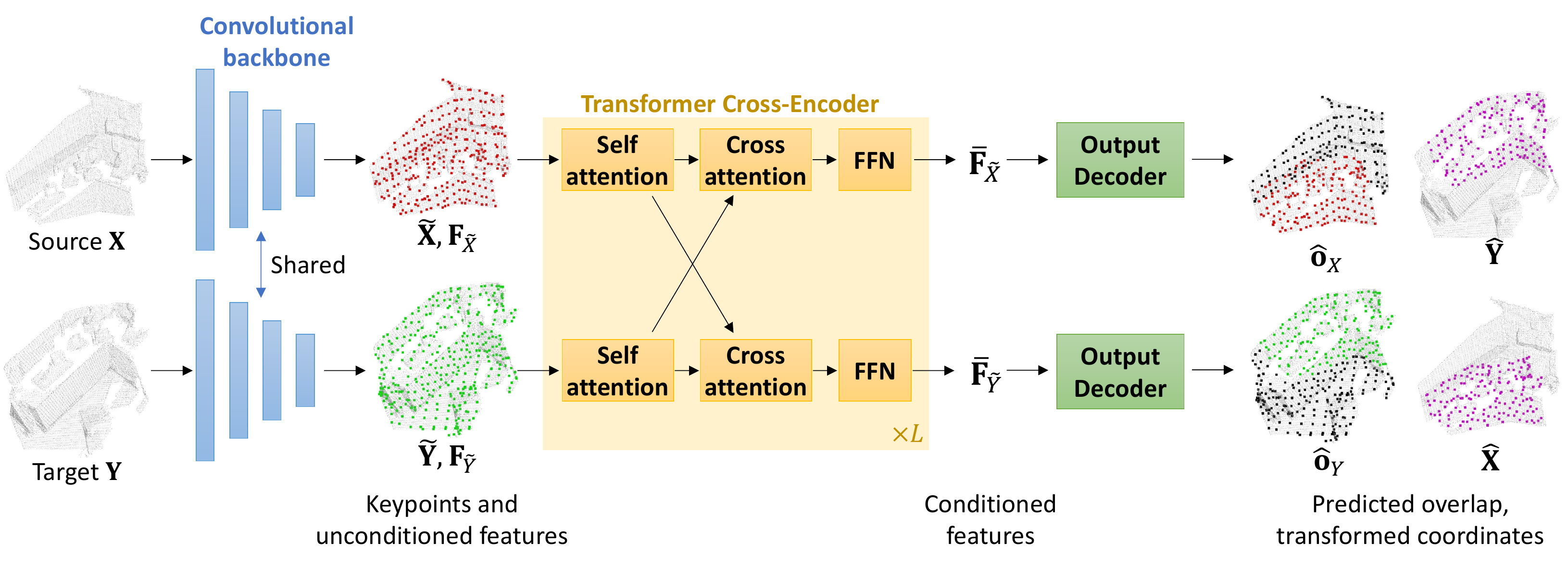}
    \caption{REGTR uses the KPConv convolutional backbone to extract a set of features for a sparse set of points. The features are then passed into several transformer cross-encoder layers. Lastly, the output decoder predicts the overlap score and the corresponding transformed coordinates of the sparse keypoints, which can be used for direct estimation of the pose. Best viewed in color.}
    \label{fig:network}
\end{figure*}

\vspace{-4.2mm}
\paragraph{Transformers.}
Transformers \cite{vaswani2017attention} propose a novel attention mechanism that makes use of multiple layers of self and cross multi-head attention to exchange information between the input and output. Although originally designed for NLP tasks, the attention mechanism has recently been shown to be useful for many computer vision tasks \cite{carion2020detr,misra2021-3detr,dosovitskiy2020vit,yu2021pointr}, and we utilize it in our work to predict point correspondences. 

\section{Problem Definition}
Consider two point clouds $\mathbf{X} \in \R^{M \times 3}$ and $\mathbf{Y} \in \R^{N \times 3}$, which we denote as the \emph{source} and \emph{target}, respectively. The objective of point cloud registration is to recover the unknown rigid transformation consisting of a rotation $\mathbf{R} \in \operatorname{SO}(3)$ and translation $\mathbf{t} \in \R^3$ that aligns $\mathbf{X}$ to $\mathbf{Y}$.

\section{Our Approach}
\Cref{fig:network} illustrates our overall framework.
We first convert the input point clouds into a smaller set of downsampled keypoints $\mathbf{\tilde{X}} \in \R^{M' \times 3}$ and $\mathbf{\tilde{Y}} \in \R^{N' \times 3}$ with $M' < M, N' < N$, and their associated features $\mathbf{F}_{\tilde{X}} \in \R^{M' \times D}, \mathbf{F}_{\tilde{Y}} \in \R^{N' \times D}$ (\cref{sect:feat-extract}). Our network then passes these keypoints and features into several transformer cross-encoder layers (\cref{sect:transformer-layers}) before finally outputting the corresponding transformed locations $\mathbf{\hat{Y}} \in \R^{M' \times 3}, \mathbf{\hat{X}} \in \R^{N' \times 3}$ of the keypoints in the other point cloud (\cref{sect:output-decode}).
The correspondences can then be obtained from the rows of $\mathbf{\tilde{X}}$ and $\mathbf{\hat{Y}}$, \ie $\{\mathbf{\tilde{x}}_i \leftrightarrow \mathbf{\hat{y}}_i\}$ and similarly for the other direction.
Concurrently, our network outputs overlap scores $\mathbf{\hat{o}}_X \in \R^{M' \times 1}, \mathbf{\hat{o}}_Y \in \R^{N' \times 1}$ that indicate the probability of each keypoint lying in the overlapping region. 
Finally, the required rigid transformation can be estimated directly from the correspondences within the overlap region (\cref{sect:rigid-est}). 

\subsection{Downsampling and Feature Extraction}\label{sect:feat-extract}
We follow \cite{bai2020d3feat,huang2021predator} to adopt the Kernel Point Convolution (KPConv) \cite{thomas2019kpconv} backbone for feature extraction. 
The KPConv backbone uses a series of ResNet-like blocks and strided convolutions to transform each input point cloud into a reduced set of keypoints $\mathbf{\tilde{X}} \in \R^{M' \times 3}, \mathbf{\tilde{Y}} \in \R^{N' \times 3}$ and their associated features $\mathbf{F}_{\tilde{X}} \in \R^{M' \times D}, \mathbf{F}_{\tilde{Y}} \in \R^{N' \times D}$.
In contrast to \cite{bai2020d3feat, huang2021predator} that subsequently perform upsampling to obtain feature descriptors of the original point cloud resolution, our approach directly predicts the transformed keypoint locations using the downsampled features.

\subsection{Transformer Cross-Encoder}\label{sect:transformer-layers}

The KPConv features from the previous step are linearly projected into a lower dimension $d=256$. These projected features are then fed into $L=6$ transformer cross-encoder\footnote{We name the layers cross-encoder layers to differentiate them from the usual transformer encoder layers \cite{vaswani2017attention} which only take in a single source.} layers. Each transformer cross-encoder layer has three sub-layers: 1) a multi-head self-attention layer operating on the two point clouds separately, 2) a multi-head cross-attention layer which updates the features using information from the other point cloud, and 3) a position-wise feed-forward network.
The cross-attention enables the network to compare points from the two different point clouds, and the self-attention allows points to interact with other points within the same point cloud when predicting its own transformed position, \eg using rigidity constraints.
Note that the network weights are shared among the two point clouds but not among the layers.

\vspace{-3mm}
\paragraph{Attention sub-layers.} The multi-head attention \cite{vaswani2017attention} operation in each sub-layer is defined as follows:
\begin{subequations}\label{eq:multihead}
\begin{equation}
    \operatorname{MHAttn}(\mathbf{Q}, \mathbf{K}, \mathbf{V}) = 
    \left
        (\text{Head}_1 \oplus ... \oplus \text{Head}_H
    \right) \mathbf{W}^O
\end{equation}
\vspace{-1mm}
\begin{equation}
    \text{Head}_{h} = \operatorname{Attn}\left(
    \mathbf{Q}\mathbf{W}_h^Q, \mathbf{K}\mathbf{W}_h^K, \mathbf{V}\mathbf{W}_h^V
    \right),
\end{equation}
\end{subequations}
where $\oplus$ denotes concatenation over the channel dimension, $\mathbf{W}_h^Q, \mathbf{W}_h^K, \mathbf{W}_h^V \in \R^{d \times d_\text{head}}$ and $\mathbf{W}^O \in \R^{Hd_\text{head} \times d}$ are learned projection matrices. We set the number of heads $H$ to 8, and $d_\text{head} = d/H$. Each attention head employs a single-head dot product attention:
\begin{equation}\label{eq:singlehead}
    \operatorname{Attn}(\mathbf{Q}, \mathbf{K}, \mathbf{V}) = \operatorname{softmax}\left(
    \frac{\mathbf{Q}\mathbf{K}^{\top}}{\sqrt{d_\text{head}}}
    \right)
    \mathbf{V}.
\end{equation}
Residual connections and layer normalization are applied to each sub-layer, and we use the ``pre-LN'' \cite{xiong2020prenorm} ordering which we find to be easier to optimize.

The query, key, values are set to the same point cloud in the self-attention layers, \ie $\operatorname{MHAttn}(\mathbf{F}_{\tilde{X}}, \mathbf{F}_{\tilde{X}}, \mathbf{F}_{\tilde{X}})$ for the source point cloud (and likewise for the target point cloud). This allows points to attend to other parts within the same point cloud.
For the cross-attention layers, the keys and values are set to be the features from the other point cloud, \ie $\operatorname{MHAttn}(\mathbf{F}_{\tilde{X}}, \mathbf{F}_{\tilde{Y}}, \mathbf{F}_{\tilde{Y}})$ for the source point cloud (and likewise for the target point cloud) to allow each point to interact with points in the other point cloud.

\vspace{-4mm}
\paragraph{Position-wise Feed-forward Network.} 
This sub-layer operates on the features of each keypoint individually. Following its usual implementation \cite{vaswani2017attention}, we use a two-layer feed-forward network with a ReLU activation function after the first layer. Similar to the attention sub-layers, residual connections and layer normalization are applied.

\vspace{-4mm}
\paragraph{Positional encodings.} Unlike previous works \cite{huang2021predator,wang2019dcp} that use attentions to learn discriminative features, our transformer layers replace the role of RANSAC and thus requires information of the point positions.
Specifically, we incorporate positional information by adding sinusoidal positional encodings \cite{vaswani2017attention} to the inputs at each transformer layer.

\medskip
The outputs of the transformer cross-encoder layers are features $\mathbf{\bar{F}}_{\tilde{X}} \in \R^{M' \times d}, \mathbf{\bar{F}}_{\tilde{Y}} \in \R^{N' \times d}$ which are conditioned on the other point cloud.

\subsection{Output Decoding}\label{sect:output-decode}
The conditioned features can now be used to predict the coordinates of transformed keypoints. To this end, we use a two-layer MLP to regress the required coordinates. 
Particularly, the corresponding locations $\mathbf{\hat{Y}} \in \R^{M' \times 3}$ of the source keypoints $\mathbf{\tilde{X}}$ in the target point cloud are given as:
\begin{equation}\label{eq:regress-coor}
    \mathbf{\hat{Y}} = \operatorname{ReLU}(\mathbf{\bar{F}}_{\tilde{X}} \mathbf{W}_1 + \mathbf{b}_1) \mathbf{W}_2 + \mathbf{b}_2,
\end{equation}
where $\mathbf{W}_1, \mathbf{W}_2$ and $\mathbf{b}_1, \mathbf{b}_2$ are learnable weights and biases, respectively. We use the hat accent $\hat{(\cdot)}$ to indicate predicted quantities. A similar procedure is used to obtain the predicted transformed locations $\mathbf{\hat{X}}$ of the target keypoints.

Alternatively, we also explore the use of a single-head attention layer (\cf \cref{table:loss-ablation}), where the predicted locations $\mathbf{\hat{Y}}$ are a weighted sum of the target keypoint coordinates $\mathbf{\tilde{Y}}$:
\begin{equation}\label{eq:weighted-coor}
    \mathbf{\hat{Y}} = \operatorname{Attn}(\mathbf{\bar{F}}_X \mathbf{W}_{\text{out}}^Q, \mathbf{\bar{F}}_Y \mathbf{W}_{\text{out}}^K, \mathbf{\tilde{Y}}),
\end{equation}
where $\operatorname{Attn}(\cdot)$ is defined previously in \cref{eq:singlehead}, and
$\mathbf{W}_{\text{out}}^Q, \mathbf{W}_{\text{out}}^K \in \R^{d \times d}$ are learned projection matrices.

In parallel, we separately predict the overlap confidence $\mathbf{\hat{o}}_X \in \R^{M' \times 1}, \mathbf{\hat{o}}_Y \in \R^{N' \times 1}$ using a single fully connected layer with sigmoid activation.
This is used to mask out the influence of correspondences outside the overlap region which are not predicted as accurately.

\subsection{Estimation of Rigid Transformation}\label{sect:rigid-est}
The predicted transformed locations in both directions are first concatenated to obtain the final set of $M' + N'$ correspondences:
\begin{equation}
    \mathbf{\hat{X}}_{\text{corr}} = 
    \begin{bmatrix}\mathbf{\tilde{X}} \\ \mathbf{\hat{X}} \\ \end{bmatrix},
    \enskip
    \mathbf{\hat{Y}}_{\text{corr}} = 
    \begin{bmatrix}\mathbf{\hat{Y}} \\ \mathbf{\tilde{Y}} \\ \end{bmatrix},
    \enskip
    \mathbf{\hat{o}}_{\text{corr}} = 
    \begin{bmatrix}\mathbf{\hat{o}}_X \\ \mathbf{\hat{o}}_Y \end{bmatrix}.
\end{equation}
The required rigid transformation can be estimated from the estimated correspondences by solving the following:
\begin{equation}
    \mathbf{\hat{R}, \hat{t}} = \argmin_{\mathbf{R}, \mathbf{t}} 
      \sum_i^{M'+N'}{\hat{o}_i \norm{\mathbf{R} \mathbf{\hat{x}}_i + \mathbf{t} - \mathbf{\hat{y}}_i}}^2,
\label{eq:rigidTransform}
\end{equation}
where $\mathbf{\hat{x}}_i, \mathbf{\hat{y}}_i, \hat{o}_i$ denote the $i^{\text{th}}$ row of $\mathbf{\hat{X}}_{\text{corr}}, \mathbf{\hat{Y}}_{\text{corr}}, \mathbf{\hat{o}}_{\text{corr}}$, \mbox{respectively}.
We follow \cite{gojcic2020multiview,yew2020rpmnet} to solve \cref{eq:rigidTransform} in closed form using a weighted variant of the Kabsch-Umeyama \cite{kabsch1976svd,umeyama1991svd} algorithm.

\subsection{Loss Functions}
\begin{figure}[t]
    \centering
    \includegraphics[width=0.99\linewidth]{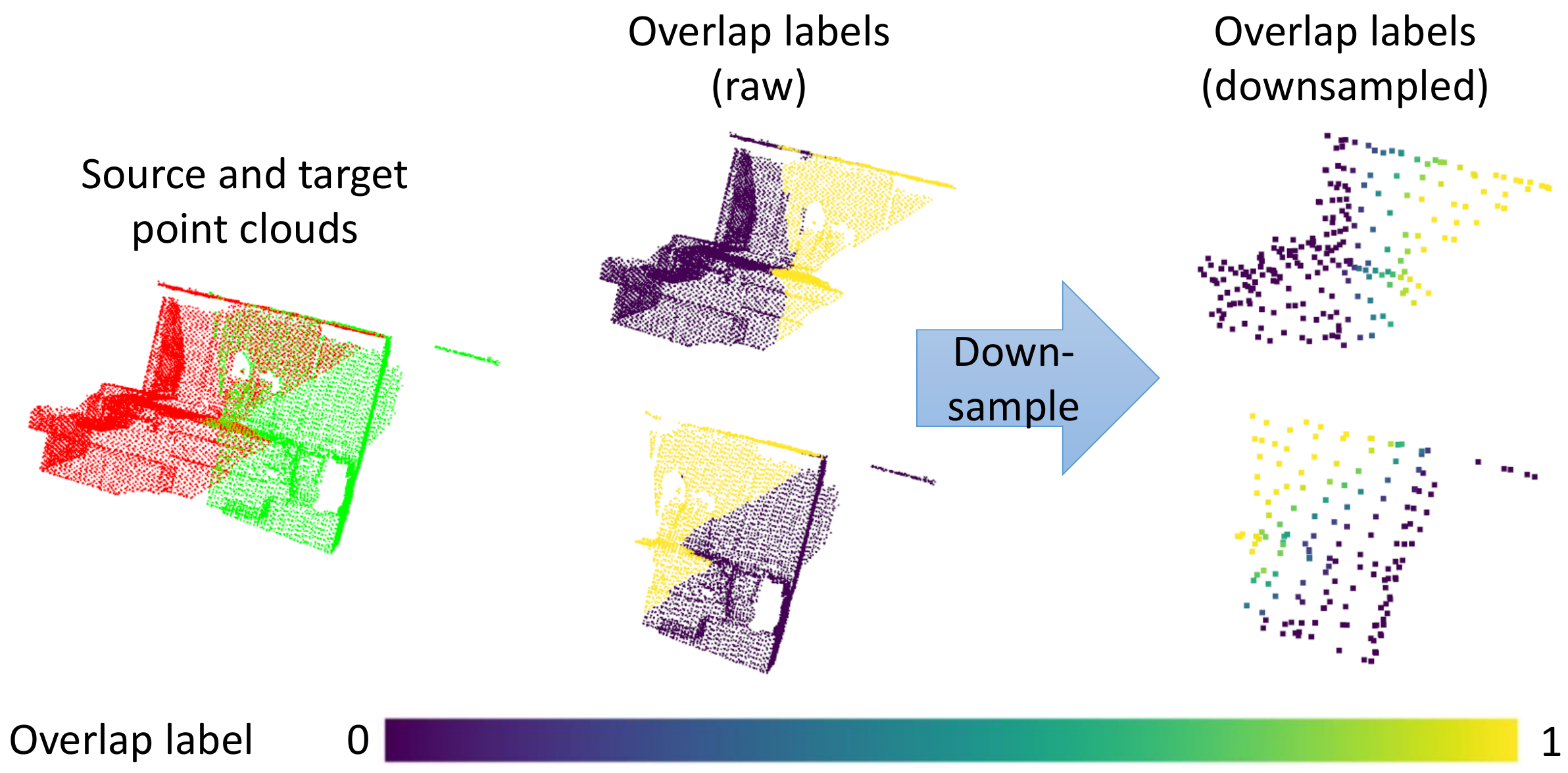}
    \caption{Pair of point clouds (left) and their corresponding ground truth overlap labels for the dense points (middle) and downsampled keypoints (right). Note that the keypoints near the overlap boundaries have a ground truth label between 0 and 1.}
    \label{fig:overlap-labels}
\end{figure}

We train our network end-to-end with the ground truth poses $\{\mathbf{R}^{*}, \mathbf{t}^{*}\}$ for supervision using the following losses:

\vspace{-4mm}
\paragraph{Overlap loss.} The predicted overlap scores are supervised using the binary cross entropy loss. The loss for the source point cloud $\mathbf{X}$ is given by:
\begin{equation}
    \mathcal{L}_{o}^{X} = -\frac{1}{M'}\sum_{i}^{M'}{
        o_{\mathbf{\tilde{x}}_i}^* \cdot \log \hat{o}_{\mathbf{\tilde{x}}_i} + 
        (1 - o_{\mathbf{\tilde{x}}_i}^*) \cdot \log \left(1 - \hat{o}_{\mathbf{\tilde{x}}_i}\right)
    }.
\end{equation}
To obtain the ground truth overlap labels $o_{\mathbf{\tilde{x}}_i}^*$, we first compute the dense ground truth labels for the original point cloud in a similar fashion as \cite{huang2021predator}.
Specifically, the ground truth label for point $\mathbf{x}_i \in \mathbf{X}$ is defined as:
\begin{equation}
    o_{\mathbf{x}_i}^* = 
    \begin{cases}
    1, & \norm{\mathcal{T}^{*}(\mathbf{x}_i) - \text{NN}(\mathcal{T}^{*}(\mathbf{x}_i), \mathbf{Y})} < r_o \\
    0, & \text{otherwise}
    \end{cases},
\end{equation}
where $\mathcal{T}^{*}(\mathbf{x}_i)$ denotes the application of the ground truth rigid transform $\{\mathbf{R}^{*}, \mathbf{t}^{*}\}$, $\text{NN}(\cdot)$ denotes the spatial nearest neighbor and $r_o$ is a predefined overlap threshold. 
We then obtain the overlap labels $o_{\mathbf{\tilde{x}}_i}^{*}$ for the downsampled keypoints through average pooling using the same pooling indices from the KPConv downsampling. 
See \cref{fig:overlap-labels} for an example of our overlap ground truth labels.
The loss $\mathcal{L}_{o}^{Y}$ for the target point cloud $\mathbf{Y}$ is obtained in a similar fashion. We thus get a total overlap loss of: $\mathcal{L}_{o} = \mathcal{L}_{o}^{X} + \mathcal{L}_{o}^{Y}$.

\vspace{-4mm}
\paragraph{Correspondence loss.} We apply a $\ell^1$ loss on the predicted transformed locations for keypoints in the overlapping region:
\begin{equation}
    \mathcal{L}_{c}^{X} =
    \frac{1}{\sum_{i}{o^*_{\mathbf{\tilde{x}}_i}}}
    {\sum_{i}^{M'}
    o_{\mathbf{\tilde{x}}_i}^* \lvert
    \mathcal{T}^*(\mathbf{\tilde{x}}_i) - \mathbf{\hat{y}}_i
    \rvert,}
\end{equation}
and similarly for the target point cloud. We thus get a total correspondence loss of: $\mathcal{L}_{c} = \mathcal{L}_{c}^{X} + \mathcal{L}_{c}^{Y}$.

\vspace{-4mm}
\paragraph{Feature loss.}
To encourage the network to take into account geometric properties when computing the correspondences, we apply an InfoNCE \cite{oord2018infonce} loss on the conditioned features. Considering the set of points $\mathbf{x} \in \mathbf{\tilde{X}}$ with a correspondence in $\mathbf{\tilde{Y}}$, the InfoNCE loss for the source point cloud is:
\begin{equation}\label{eq:infonce}
    \mathcal{L}_{f}^{X} = - \E_{\mathbf{x} \in \mathbf{\tilde{X}}} \left[
    \log
    \frac{f(\mathbf{x}, \mathbf{p_x})}
    {f(\mathbf{x}, \mathbf{p_x}) + \sum_{\mathbf{n_x}} f(\mathbf{x}, \mathbf{n_x})}
    \right],
\end{equation}
where we follow \cite{oord2018infonce} to use a log-bilinear model for $f(\cdot,\cdot)$:
\begin{equation}
    f(\mathbf{x}, \mathbf{c}) = \exp{\left( 
    {\mathbf{\bar{f}}_\mathbf{x}}^T 
    \mathbf{W}_{f} 
    {\mathbf{\bar{f}}_\mathbf{c}}
    \right)}.
\end{equation}
${\mathbf{\bar{f}}_\mathbf{x}}$ denotes the conditioned feature for point $\mathbf{x}$. $\mathbf{p_x}$ and $\mathbf{n_x}$ denote keypoints in $\mathbf{\tilde{Y}}$ which match and do not match $\mathbf{x}$, respectively. They are determined using the positive and negative margins $(r_p, r_n)$ which are set as $(m, 2m)$. $m$ is the voxel distance used in the final downsampling layer in the KPConv backbone, and all negative points that falls outside the negative margin are utilized for $\mathbf{n_x}$. 
Since the two point clouds contain the same type of features, we enforce the learnable linear transformation $\mathbf{W}_f$ to be symmetrical by parameterizing it as the sum of a upper triangular matrix $\mathbf{U}_f$ and its transpose, \ie $\mathbf{W}_f = \mathbf{U}_f + \mathbf{U}_f^{\top}$.
As explained in \cite{oord2018infonce}, \cref{eq:infonce} maximizes the mutual information between features for matching points. Unlike \cite{bai2020d3feat,huang2021predator}, we do not use the circle loss \cite{sun2020circle} that requires the matching features to be similar (w.r.t. $\ell^2$ or cosine distance). This is unsuitable since: 1) our conditioned features contains information about the transformed positions, and 2) our keypoints are sparse and are unlikely to be at the same location in the two point clouds, and therefore the geometric features are also different. 

\medskip
Our final loss is a weighted sum of the three components: $\mathcal{L} = \mathcal{L}_c + \lambda_o \mathcal{L}_o + \lambda_f \mathcal{L}_f$, where we set $\lambda_o=1.0$ and $\lambda_f=0.1$ for all experiments.

\section{Experiments}
\subsection{Implementation Details}
We train our network using AdamW \cite{loshchilov2019adamw} optimizer with a initial learning rate of 0.0001 and weight decay of 0.0001. Gradients are clipped at 0.1. For the 3DMatch dataset, we train for 60 epochs with a batch size of 2, halving the learning rate every 20 epochs. We train on the ModelNet40 dataset for 400 epochs with a batch size of 4, and halving the learning rate every 100 epochs. Training requires around 2.5 and 2 days for 3DMatch and ModelNet40 on a single Nvidia Titan RTX, respectively.

\subsection{Datasets and Results}

\paragraph{3DMatch.}
The 3DMatch dataset \cite{zeng20163dmatch} contains 46 train, 8 validation and 8 test scenes. We use the preprocessed data from \cite{huang2021predator} containing voxel-grid downsampled point clouds, and follow them to evaluate on both pairs with $>30\%$ overlap (\emph{3DMatch}) and $10-30\%$ overlap (\emph{3DLoMatch}). Each input point cloud contains an average of about 20,000 points, which are downsampled to an average of 345 points by our KPConv backbone. We perform training data augmentation by applying small rigid perturbations, jittering of the point locations and shuffling of points.

Following \cite{choi2015robustrecon,huang2021predator,bai2020d3feat}, we evaluate using Registration Recall (RR) which measures the fraction of successfully registered pairs, defined as having a correspondence RMSE below 0.2m. We also evaluate on the Relative Rotation Errors (RRE) and Relative Translation Errors (RTE) that measures the accuracy of successful registrations.
We follow \cite{huang2021predator} and compare against several recent learned correspondence-based algorithms \cite{gojcic2019perfect,choy2019fcgf,bai2020d3feat,huang2021predator}\footnote{The initial Predator code had a \href{https://github.com/overlappredator/OverlapPredator/issues/15}{bug} which decreased the performance, and we list the improved results using its corrected version.}. These algorithms tend to perform better with a larger number of sampled interest points, and therefore we only show the results for the maximum number (5000) of sampled points. Since Predator \cite{huang2021predator} obtains the highest registration recall when using 1000 interest points, we also reran their open-source code with 1000 interest points and include the results under Predator-1k.
Furthermore, we compare with several methods \cite{xu2021omnet,choy2020dgr,cao2021pcam} designed to avoid RANSAC. We trained OMNet \cite{xu2021omnet} on 3DMatch with a batch size of 32 for 2000 epochs using 1024 random points. For \cite{choy2020dgr,cao2021pcam}, we use the authors' trained weights. We disabled ICP refinement in DGR \cite{choy2020dgr} for a fair comparison.

\begin{figure*}[t]
    \includegraphics[width=\linewidth]{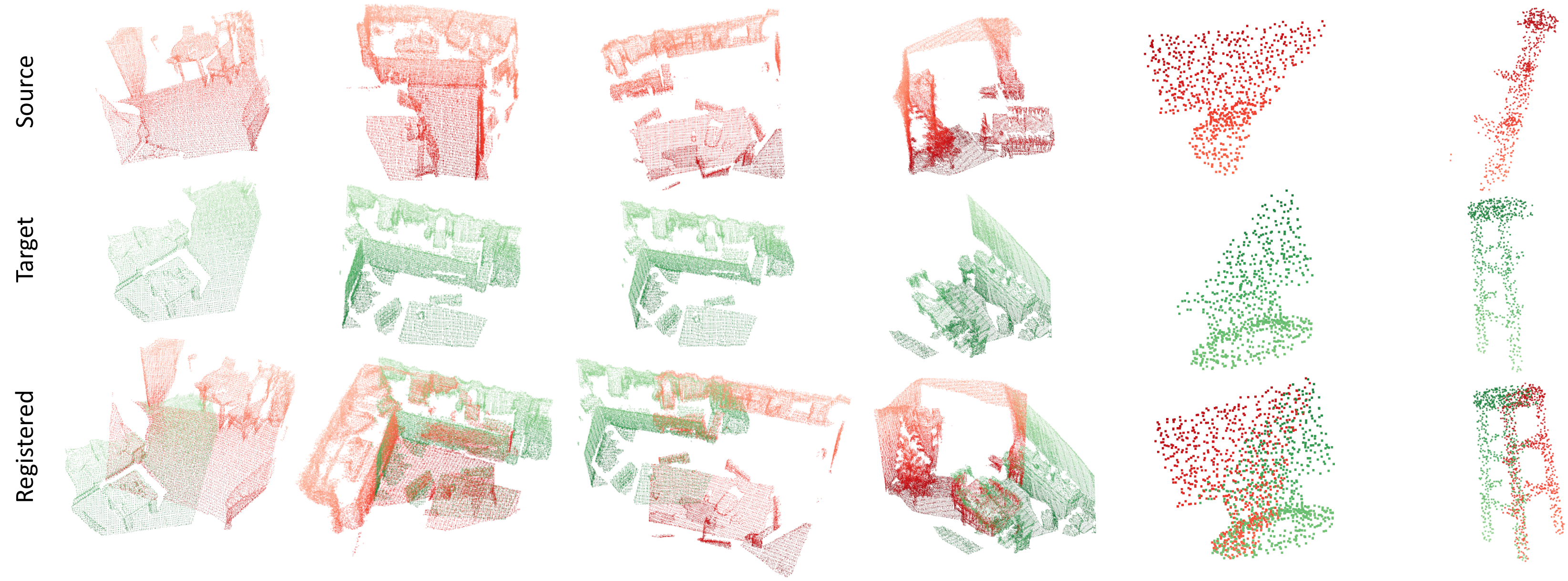}
    \vspace{-6mm}
    \\
    \begin{subfigure}{0.21\linewidth}\caption{}\label{fig:qualitative-results-a}\end{subfigure}
    \begin{subfigure}{0.13\linewidth}\caption{}\label{fig:qualitative-results-b}\end{subfigure}
    \begin{subfigure}{0.18\linewidth}\caption{}\label{fig:qualitative-results-c}\end{subfigure}
    \begin{subfigure}{0.18\linewidth}\caption{}\label{fig:qualitative-results-d}\end{subfigure}
    \begin{subfigure}{0.16\linewidth}\caption{}\label{fig:qualitative-results-e}\end{subfigure}
    \hfill
    \begin{subfigure}{0.05\linewidth}\caption{}\label{fig:qualitative-results-f}\end{subfigure}
    \vspace{-2mm}
    \caption{Example qualitative registration results for (a, b) 3DMatch, (c, d) 3DLoMatch (e) ModelNet40, and (f) ModelLoNet.}
    \label{fig:qualitative-results}
\end{figure*}

\Cref{table:3dmatch-quantitative} shows the quantitative results, and \cref{fig:teaser,fig:qualitative-results-a,fig:qualitative-results-b,fig:qualitative-results-c,fig:qualitative-results-d} show several examples of the qualitative results. We also show the results for the individual scenes in the supplementary. For both 3DMatch and 3DLoMatch benchmarks, our method achieves the highest average registration recall across scenes. Interestingly, our registration is also very precise and achieved the lowest RTE and RRE on both 3DMatch and 3DLoMatch benchmarks despite only using a small number of points for pose estimation.
In addition, we also compare with Predator-NR, a RANSAC-free variant of Predator-1k that utilizes the product of the predicted overlap and matchability scores to weigh the correspondences during pose estimation. The underperformance of Predator-NR indicates that our proposed method is more suitable for replacing RANSAC.
The results support our claim that our attention mechanism can replace the role of RANSAC since our REGTR uses largely the same KPConv backbone as Predator.
Lastly, we note that OMNet does not perform well on the 3DMatch dataset. This is likely due to the difficulty in describing complex scenes with a single global feature vector. This behavior is also previously observed in \cite{choy2020dgr} for another global feature-based algorithm, PointNetLK \cite{aoki2019pointnetlk}.

\begin{table}
\footnotesize
\centering
\setlength\tabcolsep{2.2pt}
\begin{tabularx}{\linewidth}{X | c c c | c c c}
  \hline
  & \multicolumn{3}{c|}{\emph{3DMatch}} & \multicolumn{3}{c}{\emph{3DLoMatch}} \\
  Method & RR(\%) & RRE(\degree) & RTE(m)
  & RR(\%) & RRE(\degree) & RTE(m)
  \\
  \hline
  3DSN \cite{gojcic2019perfect} & 78.4 & 2.199 & 0.071 & 33.0 & 3.528 & 0.103
  \\
  FCGF \cite{choy2019fcgf} & 85.1 & 1.949 & 0.066 & 40.1 & 3.147 & 0.100
  \\
  D3Feat \cite{bai2020d3feat} & 81.6 & 2.161 & 0.067 & 37.2 & 3.361 & 0.103
  \\
  Predator-5k \cite{huang2021predator} & 89.0 & 2.029 & 0.064 & 59.8 & \underline{3.048} & \underline{0.093}
  \\
  Predator-1k \cite{huang2021predator} & \underline{90.5} & 2.062 & 0.068 & \underline{62.5} & 3.159 & 0.096
  \\
  Predator-NR \cite{huang2021predator} & 62.7 & 2.582 & 0.075 & 24.0 & 5.886 & 0.148
  \\
  \hline
  OMNet \cite{xu2021omnet} & 35.9 & 4.166 & 0.105 & 8.4 & 7.299 & 0.151
  \\
  DGR \cite{choy2020dgr} & 85.3 & 2.103 & 0.067 & 48.7 & 3.954 & 0.113
  \\
  PCAM \cite{cao2021pcam} & 85.5 & \underline{1.808} & \underline{0.059} & 54.9 & 3.529 & 0.099
  \\
  \hline
  Ours & \textbf{92.0} & \textbf{1.567} & \textbf{0.049} & \textbf{64.8} & \textbf{2.827} & \textbf{0.077} \\
  \hline
\end{tabularx}
\vspace{-1mm}
\caption{Performance on 3DMatch and 3DLoMatch datasets. Results for 3DSN, FCGF, D3Feat and Predator-5k are from \cite{huang2021predator}.}
\label{table:3dmatch-quantitative}
\end{table}

\begin{table}
\footnotesize
\centering
\setlength\tabcolsep{3.5pt}
\begin{tabularx}{\linewidth}{X | c c c | c c c}
  \hline
  & \multicolumn{3}{c|}{\emph{ModelNet}} & \multicolumn{3}{c}{\emph{ModelLoNet}} \\
  Methods & RRE & RTE & CD & RRE & RTE & CD \\
  \hline
  PointNetLK \cite{aoki2019pointnetlk} &
  29.725 & 0.297 & 0.0235 & 48.567 & 0.507 & 0.0367
  \\
  OMNet \cite{xu2021omnet} &
    2.947 & 0.032 & 0.0015 & 6.517 & 0.129 & 0.0074  
  \\
  DCP-v2 \cite{wang2019dcp} 
  & 11.975 & 0.171 & 0.0117 
  & 16.501 & 0.300 & 0.0268
  \\
  RPM-Net \cite{yew2020rpmnet}
  & \underline{1.712} & \underline{0.018} & \underline{0.00085} 
  & 7.342 & \underline{0.124} & \underline{0.0050} 
  \\
  Predator \cite{huang2021predator} 
  & 1.739 & 0.019 & 0.00089
  & \underline{5.235} & 0.132 & 0.0083
  \\
  \hline
  Ours & \textbf{1.473} & \textbf{0.014} & \textbf{0.00078} 
  & \textbf{3.930} & \textbf{0.087} & \textbf{0.0037}\\
  \hline
\end{tabularx}
\vspace{-1mm}
\caption{Evaluation results on ModelNet40 dataset. Results of DCP-v2, RPM-Net and Predator are taken are from \cite{huang2021predator}.}
\label{table:modelnet-performance}
\end{table}

\vspace{-4mm}
\paragraph{ModelNet40.}
We also evaluate on the ModelNet40 \cite{wu2015modelnet} dataset comprising synthetic CAD models. We follow the data setting in \cite{yew2020rpmnet,huang2021predator}, where the point clouds are sampled randomly from mesh faces of the CAD models, cropped and subsampled. Following \cite{huang2021predator}, we evaluate on two partial overlap settings: \emph{ModelNet} which has 73.5\% pairwise overlap on average, and \emph{ModelLoNet} which contains a lower 53.6\% average overlap.  We train only on ModelNet, and perform direct generalization to ModelLoNet.
We follow \cite{yew2020rpmnet,huang2021predator} and measure the performance using Relative Rotation Error (RRE) and Relative Translation Error (RTE) on all point clouds, as well as the Chamfer distance (CD) between the registered scans.

The results are shown in \cref{table:modelnet-performance}, with example qualitative results in \cref{fig:qualitative-results-e,fig:qualitative-results-f}.
We compare against recent correspondence-based \cite{huang2021predator} and end-to-end registration methods \cite{wang2019dcp,yew2020rpmnet,aoki2019pointnetlk,xu2021omnet}. Predator \cite{huang2021predator} samples 450 points in this experiment. 
OMNet \cite{xu2021omnet} was originally trained only on axis-asymmetrical categories, and we retrained it on all categories to obtain a slightly improved result.
As noted in \cite{huang2021predator}, many of the end-to-end registration methods are specifically tuned for ModelNet. RPM-Net \cite{yew2020rpmnet} additionally uses surface normal information. Despite this, our REGTR substantially outperforms all baseline methods in all metrics under both normal overlap (ModelNet) and low overlap (ModelLoNet) regimes. Our learned attention mechanism is able to outperform the optimal transport (in RPM-Net) and RANSAC step (in Predator).

\begin{table}[t]
    \footnotesize
    \centering
    \begin{tabularx}{\linewidth}{X | c c c | c}
    \hline
    Method & Preproc. & Feat. Extract. & Pose est. & Total \\
    \hline
    3DSN \cite{gojcic2019perfect} & 27938 & 1872 & 2588 & 32398
    \\
    FCGF \cite{choy2019fcgf} & 15 & 41 & 1597 & 1653
    \\
    D3Feat \cite{bai2020d3feat} & 174\textsuperscript{*} & 27 & 795 & 996 \\
    Predator-5k \cite{huang2021predator} & 245\textsuperscript{*} & 45 & 1017 & 1306
    \\
    Predator-1k \cite{huang2021predator} & 245\textsuperscript{*} & 45 & 189 & 479
    \\
    \hline
    OMNet \cite{xu2021omnet} & --- & 9 & 1 & \textbf{10}
    \\
    DGR \cite{choy2020dgr} & 28 & 31 & 1258 & 1318
    \\
    PCAM \cite{cao2021pcam} & --- & 520 & 1063 & 1584
    \\
    \hline
    Ours & 35\textsuperscript{*} & 54 & 2 & \underline{91} \\
    \hline
    \end{tabularx}
    \caption{Run time in milliseconds on the 3DMatch benchmark test set. \textsuperscript{*}Similar pre-processing is used for D3Feat, Predator and our algorithm, but our algorithm uses a faster GPU implementation.}
    \label{table:timings}
\end{table}

\begin{figure}[t]
    \centering
    \includegraphics[width=0.85\linewidth,trim={0 1mm 0 0},clip]{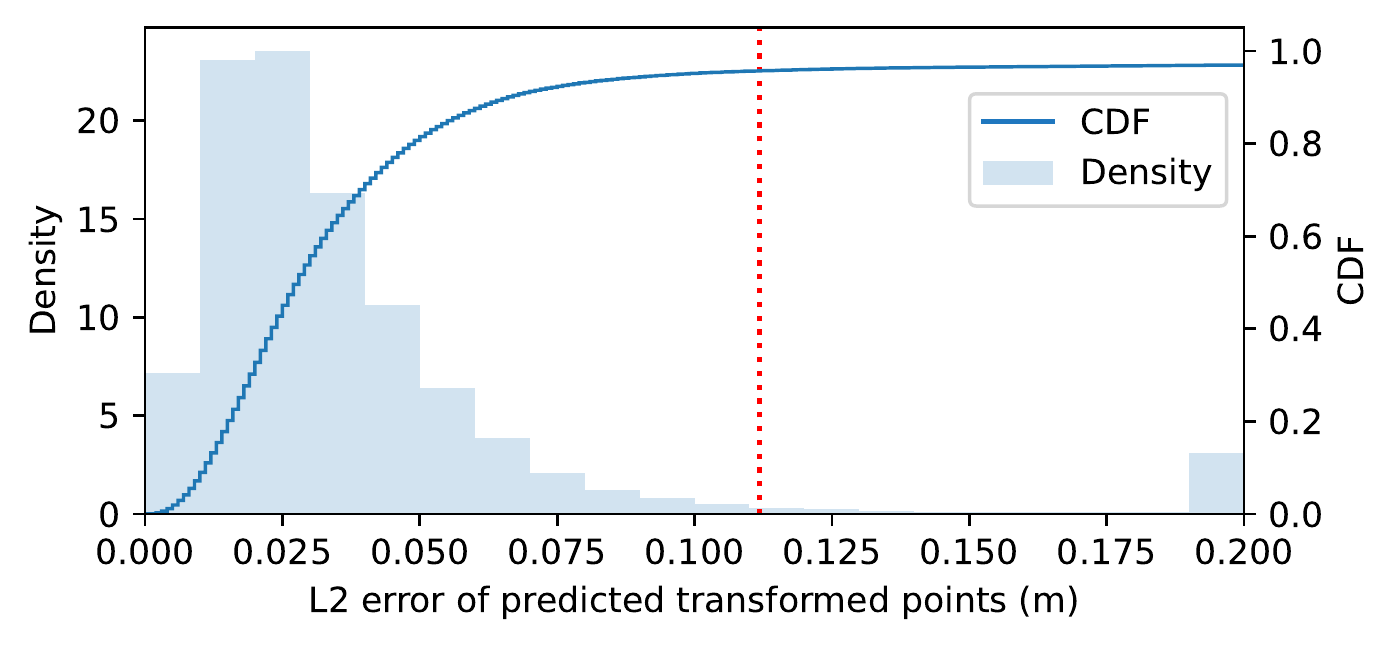}
    \vspace{-2mm}
    \caption{Histogram and CDF plot of $\ell^2$ errors of predicted correspondences. 95.7\% of predicted correspondences have errors below 0.112m, the median keypoint-to-keypoint distance (denoted by red dashed line). Errors are clipped at 0.2m for clarity, with only 3.0\% of predicted correspondences exceeding this error.}
    \label{fig:corr_err_hist}
\end{figure}

\subsection{Analysis}
We perform further analysis in this section to better understand our algorithm behavior. All experiments in this section are performed on the larger 3DMatch dataset.

\vspace{-4mm}
\paragraph{Runtime.} We compare the runtime of REGTR against several algorithms in \cref{table:timings}. We conducted the test on a single Nvidia Titan RTX with Intel Core i7-6950X @ 3.0GHz and 64GB RAM. Our entire pipeline runs under 100ms, and is feasible for many real-time applications.
The time consuming step for correspondence-based algorithms is the pose estimation which includes feature matching and RANSAC. For example, excluding preprocessing required for the KPConv backbone, Predator \cite{huang2021predator} takes 234ms for the registration when sampling just 1,000 points. DGR \cite{choy2020dgr} and PCAM \cite{cao2021pcam} also require long times for pose estimation due to their robust refinement and RANSAC safeguard. Although the global feature-based OMNet runs faster than our algorithm, we note that it is unable to obtain good accuracy on the 3DMatch dataset.

\vspace{-4mm}
\paragraph{Accuracy of predicted correspondences.} In \cref{fig:corr_err_hist}, we plot the distribution of $\ell^2$-error of the predicted correspondences for keypoints within the overlap region (where $o_{\mathbf{\tilde{x}}_i}^{*}, o_{\mathbf{\tilde{y}}_i}^{*} > 0.5)$ for the 3DMatch test set. The median error of our predicted correspondences is 0.028m, which is significantly smaller than the median distance between keypoints (0.112m). For comparison, an oracle matcher that matches every keypoint to the closest keypoint using the ground truth pose obtains a median error of 0.071m. Our direct prediction of correspondences is able to overcome the resolution issues from the downsampling, and thus explains the precise registration obtained by REGTR.

\begin{figure}[t]
    \begin{subfigure}[b]{0.421\linewidth}
        \centering
        \includegraphics[height=2.9cm]{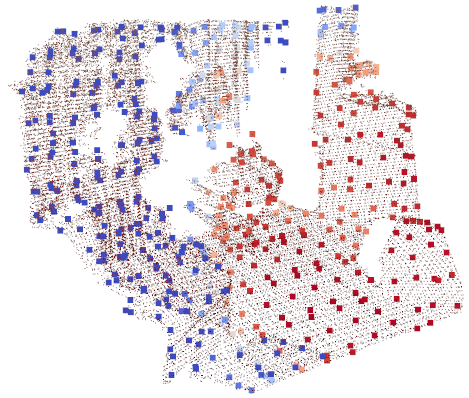}
        \vspace{-1mm}
        \caption{}
        \label{fig:viz-corr-a}
    \end{subfigure}
    \hfill
    \begin{subfigure}[b]{0.559\linewidth}
        \centering
        \includegraphics[height=2.9cm]{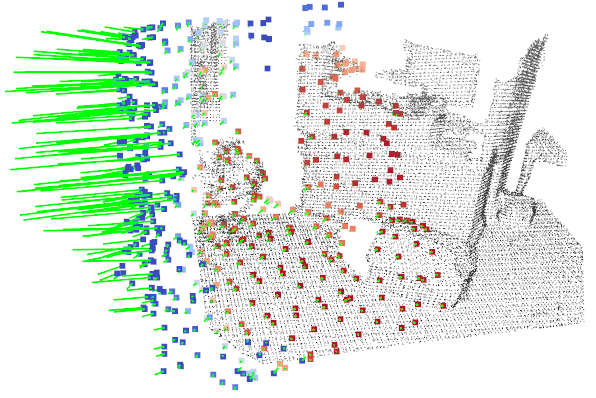}
        \vspace{-1mm}
        \caption{}
        \label{fig:viz-corr-b}
    \end{subfigure}
    \vspace{-2mm}
    \caption{Visualization of predicted correspondences. Keypoints are colored based on their predicted overlap score, where high scores are denoted in red. (a) Source $\mathbf{X}$ and keypoints $\mathbf{\tilde{X}}$, (b) Target $\mathbf{Y}$ and predicted correspondences $\mathbf{\hat{Y}}$, with green lines showing the correspondence error. Best viewed in color.}
    \label{fig:viz-corr}
\end{figure}

We also visualize the predicted correspondences of a point cloud pair in \cref{fig:viz-corr}. The short green error lines in \cref{fig:viz-corr-b} indicate our predicted transformed locations are highly accurate within the overlap region, even in non-informative regions (\eg floor).
Interestingly, correspondence for points outside the overlap region are projected to near the overlap boundaries. These observations suggest that REGTR is able to make use of rigidity constraints to guide the positions of the predicted correspondences.

\begin{figure}[t]
    \centering
    \begin{subfigure}[b]{0.32\linewidth}
        \centering
        \includegraphics[width=\textwidth]{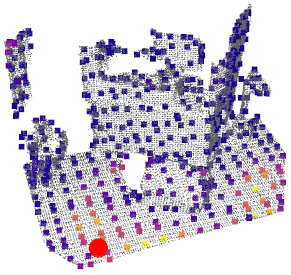}
        \caption{Cross att. (layer 1)}
        \label{fig:att_viz_cross1}
    \end{subfigure}
    \hfill
    \begin{subfigure}[b]{0.32\linewidth}
        \centering
        \includegraphics[width=\textwidth]{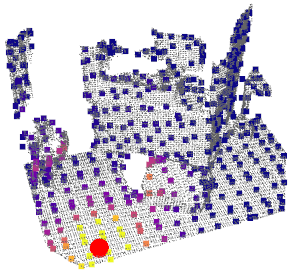}
        \caption{Cross att. (layer 6)}
        \label{fig:att_viz_cross6}
    \end{subfigure}
    \hfill
    \begin{subfigure}[b]{0.32\linewidth}
        \centering
        \includegraphics[width=\textwidth]{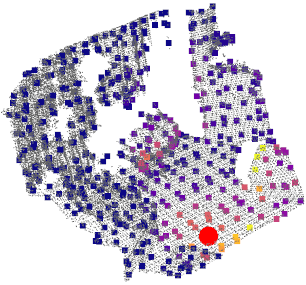}
        \caption{Self att. (layer 6)}
        \label{fig:att_viz_self6}
    \end{subfigure}
    \vspace{-2mm}
    \caption[Visualization of attention weights]{Visualization of attention weights for the point indicated with a red dot. Brighter colors indicate higher attention.}
    \label{fig:att_viz}
\end{figure}

\vspace{-4mm}
\paragraph{Visualization of attention.}
In \cref{fig:att_viz}, we visualize the attention for a point on the ground for the same point cloud pair from the previous section. Since the point lies in a non-informative region, the point attends to multiple similar looking regions in the other point cloud in the first transformer layer (\cref{fig:att_viz_cross1}). At the sixth layer, the point is confident of its position and mostly focuses on its correct corresponding location (\cref{fig:att_viz_cross6}). The self-attention in \cref{fig:att_viz_self6} shows that the point makes use of feature rich regions within the same point cloud to help localize its correct location.

\subsection{Ablations}\label{sect:ablations}
We further perform ablation studies on the 3DMatch dataset to understand the role of various components.

\begin{figure}[t]
    \centering
    \includegraphics[width=0.9\linewidth,trim={0 3.5mm 0 0},clip]{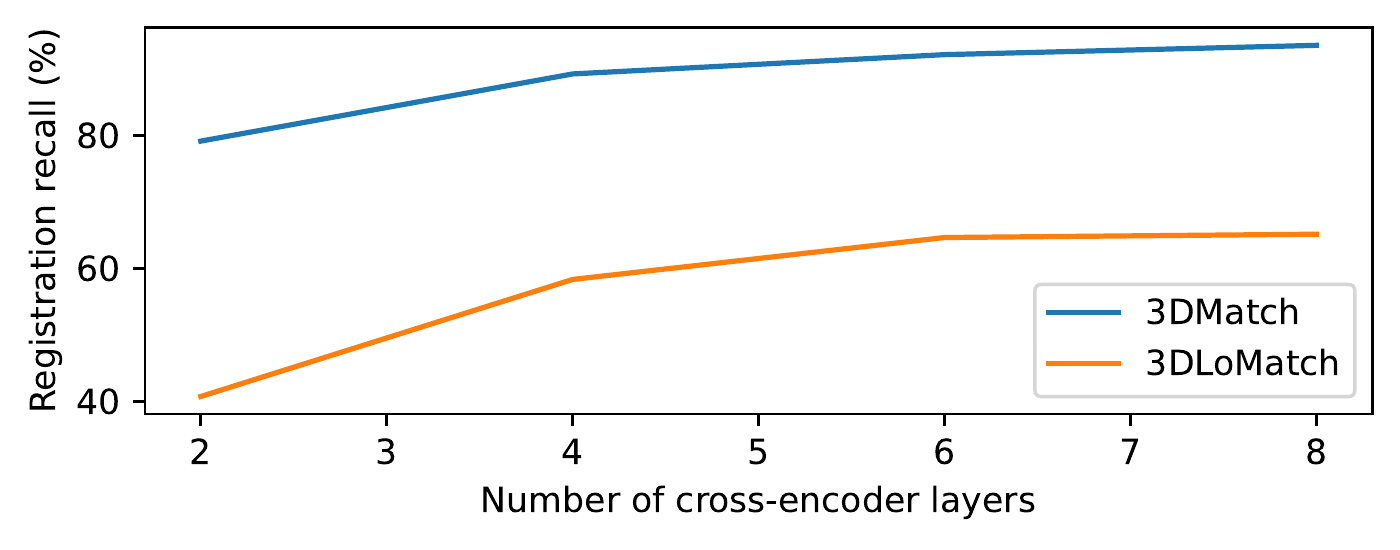}
    \vspace{-2mm}
    \caption{Performance for various number of cross-encoder layers.}
    \label{fig:num-layers}
\end{figure}

\vspace{-4mm}
\paragraph{Number of cross-encoder layers.}
We evaluate how the performance varies with the number of cross-encoder layers in \cref{fig:num-layers}. Our network cannot function without any cross-encoder layers, and we show the registration recall for two to eight layers.
Performance generally improves with more cross-encoder layers, but saturates around $L=6$ encoder layers (which we use for all our experiments).

\begin{table}[t]
\footnotesize
\centering
\setlength\tabcolsep{1.7pt}
\begin{tabularx}{\linewidth}{X | c c c | c c c}
  \hline
  & \multicolumn{3}{c|}{\emph{3DMatch}} & \multicolumn{3}{c}{\emph{3DLoMatch}} \\
  Method & RR(\%) & RRE(\degree) & RTE(m) & RR(\%) & RTE(\degree) & RTE(m) \\
  \hline
  RANSAC baseline & 87.7 & 2.296 & 0.072 & 52.7 & 4.038 & 0.112
  \\
  \hline
  Weighted coor. & 90.9 & \textbf{1.468} & \textbf{0.046} & 63.1 & \textbf{2.540} & \textbf{0.076} \\
  \hline
  No feature loss & 90.4 & 1.638 & \underline{0.049} & 61.9 & 2.898 & 0.083
  \\
  Circle loss & 90.0 & 1.696 & 0.051 & 61.2 & 3.217 & 0.092
  \\
  Loss on all layers & 83.9 & 1.781 & 0.053 & 46.8 & 3.448 & 0.101
  \\
  \hline
  REGTR & \textbf{92.0} & \underline{1.567} & \underline{0.049} & \textbf{64.8} & 2.827 & \underline{0.077}
  \\
  REGTR+RANSAC & \underline{91.9} & 1.607 & \underline{0.049} & \underline{63.3} & \underline{2.753} & 0.079
  \\
  \hline
\end{tabularx}
\vspace{-2mm}
\caption{Ablation of components and losses} \vspace{-2mm}
\label{table:loss-ablation}
\end{table}

\vspace{-4mm}
\paragraph{Comparison with RANSAC.} We compare with a version of REGTR where we replace our output decoder with a two-layer MLP which outputs 256D feature descriptors and a parallel single-layer decoder that outputs the overlap score. The pose is subsequently estimated using RANSAC on nearest neighbor feature matches. The network is trained using only $\mathcal{L}_{o}$ and $\mathcal{L}_f$ (using Circle Loss \cite{sun2020circle}). This results in a lower registration recall, and significantly higher rotation and translation errors as the downsampled keypoints do not provide enough resolution for accurate registration.

We also try applying RANSAC to the predicted correspondences from REGTR to see if the performance can further improve. Row 7 of \cref{table:loss-ablation} shows marginally worse registration recall. This indicates that RANSAC is no longer beneficial on the predicted correspondences that are already consistent with a rigid transformation.

\vspace{-4mm}
\paragraph{Decoding scheme.}
We compare with decoding the coordinates as a weighted sum of coordinates (Eq. \ref{eq:weighted-coor}). Compared to our simpler approach of regressing the coordinates using a MLP, computing the coordinates as a weighted sum achieves a slightly better RTE and RRE, but lower registration recall. See rows 2 and 6 of \cref{table:loss-ablation}.

\vspace{-4mm}
\paragraph{Loss ablations.}
Rows 3-6 of \cref{table:loss-ablation} shows the registration performance with different loss configurations. Without the feature loss to guide the network outputs, the network obtained a 1.6\% and 2.9\% lower registration recall for 3DMatch and 3DLoMatch, respectively.
Using circle loss from \cite{huang2021predator} also underperformed as the network cannot incorporate positional information into the feature as effectively.
We also experimented with applying the losses on all $L=6$ transformer layers (instead of just the final one) with the output decoder shared among all cross-encoder layers. This additional supervision led to a  8.1\% (3DMatch) and 18.0\% (3DLoMatch) lower registration recall. Consequently, we only apply the supervision for the output of the last cross-encoder layer.

\section{Limitations}
Our use of transformers layers with quadratic complexity prevents its use on large number of points, and we can only apply them on downsampled point clouds. Although our direct correspondence prediction alleviates the resolution issue, it is possible that a finer resolution can result in even higher performance. We have tried transformer layers with linear complexity \cite{katharopoulos2020lineartransformers,choromanski2021performers}, but that obtained subpar performance. Alternate workarounds include using sparse attention \cite{child2019sparsetransformers}, or performing a coarse-to-fine registration.

\section{Conclusions}
We propose the REGTR for rigid point cloud registration, which directly predicts clean point correspondences using multiple transformer layers. The rigid transformation can then be estimated from the correspondences without further nearest neighbor feature matching nor RANSAC steps.
The direct prediction of correspondences overcomes the resolution issues from the use of downsampled features, and our method achieves state-of-the-art performance on both scene and object point cloud datasets.

\vspace{-2mm}
\small
\paragraph{Acknowledgement.}
This research/project is supported in part by the National Research Foundation, Singapore under its AI Singapore Program (AISG Award No: AISG2-RP-2020-016), and the Tier 2 grant MOE-T2EP20120-0011 from the Singapore Ministry of Education.

{\small
\bibliographystyle{ieee_fullname}
\bibliography{egbib}
}

\clearpage \appendix

{\noindent \large \textbf{Supplementary Material}}

\smallskip \noindent
In this supplementary, we first provide additional details on the datasets and their preprocessing (Sec. \ref{sect:dataset-details}). We then describe the procedure for recovering the rigid transformation from the correspondences (Sec. \ref{sect:kabsch}), our sinusoidal position encodings (Sec. \ref{sect:pos-emb-details}) and additional information on the network architecture (Sec. \ref{sect:network-details}). Finally, we show detailed results for ScanNet (Sec. \ref{sect:scannet-results-details}) and additional qualitative results (Sec. \ref{sect:additional-qualitative}).

\section{Dataset Details}\label{sect:dataset-details}
\paragraph{3DMatch.}
The 3DMatch \cite{zeng20163dmatch} dataset comprises RGB-D frames obtained from several sources (\cref{table:3dmatch-datasets}). The data is captured from diverse scenes (\eg bedrooms, kitchens, offices) and different sensors (\eg Microsoft Kinect, Intel Realsense), and each point cloud is generated by fusing 50 consecutive depth frames using TSDF volumetric fusion \cite{curless1996volumetric}. We use the voxel-grid downsampled data from Predator \cite{huang2021predator}, and the same point cloud pairs for training and evaluation. The dataset contains 46 train, 8 validation, and 8 test scenes. The training and validation scenes contains a total of 20,586\footnote{one point cloud (7-scenes-fire\slash19) has a wrong groundtruth pose and we exclude training pairs containing this point cloud.} and 1,331 point cloud pairs respectively, and the test scenes contain 1,279 (3DMatch) and 1,726 (3DLoMatch) pairs. 
We apply training data augmentation by applying a small rigid perturbation with magnitudes sampled from a Gaussian distribution with $\sigma_r=0.1\pi$ and $\sigma_t=0.1$ for the rotation and translation, respectively. We then apply a Gaussian noise $(\sigma=0.05)$ on the individual point locations, and shuffling of point order.

\vspace{-2mm}
\paragraph{ModelNet40.} The ModelNet40 \cite{wu2015modelnet} dataset provides 3D CAD models from 40 object categories for academic use. We follow previous works \cite{yew2020rpmnet,wang2019dcp} in using the preprocessed data from \cite{qi2017pointnet}. These data are generated by sampling 2,048 points from the mesh faces and then scaling them to fit into a unit sphere. The partial scans are generated from the procedure in \cite{yew2020rpmnet}: A half-space with random direction is sampled, and shifted such that a proportion $p$ of points lie within the half space. Subsequently, random rotation of up to $45\degree$, translation up to 0.5 units, Gaussian noise $(\sigma=0.05)$ on the individual point locations, and shuffling of point order are applied to the point clouds. The point clouds are finally resampled to 717 points.
Following \cite{huang2021predator}, $p$ is set to 0.7 and 0.5 for ModelNet and ModelLoNet benchmarks, respectively.
We use the first 20 categories for training and validation, and the other 20 categories for testing.

\begin{table}[hbt]
\small
\centering
\begin{tabularx}{0.95\linewidth}{X | l }
    \hline
    Datasets & License \\
    \hline
    SUN3D \cite{xiao2013sun3d,halber2017fine} & CC BY-NC-SA 4.0 \\
    7-Scenes \cite{shotton2013sevenscene} & Non-commercial use only \\
    RGB-D Scenes v2 \cite{lai2014unsupervised} & (License not stated) \\
    BundleFusion \cite{dai2017bundlefusion} & CC BY-NC-SA 4.0 \\
    Analysis-by-Synthesis \cite{valentin2016analBySyn} & CC BY-NC-SA 4.0 \\
    \hline
\end{tabularx}
\vspace{-1mm}
\caption{Raw data used in the 3DMatch \cite{zeng20163dmatch} dataset and their licenses.}
\label{table:3dmatch-datasets}
\end{table}

\section{Estimation of Rigid Transformation}\label{sect:kabsch}
In this section, we describe the closed form solution for the rigid transformation $\{\mathbf{R}, \mathbf{t}\}$, given correspondences $\{\mathbf{x}_i \leftrightarrow \mathbf{y}_i\}$ with their weights $\{o_i\}$, as used in \cref{sect:rigid-est}:
\begin{equation}
    \mathbf{\hat{R}, \hat{t}} = \argmin_{\mathbf{R}, \mathbf{t}} 
      \sum_i^N{o_i \norm{\mathbf{R} \mathbf{x}_i + \mathbf{t} - \mathbf{y}_i}}^2.
\label{eq:rigidTransform-sup}
\end{equation}

\vspace{-2mm}
\paragraph{Step 1.} Compute the weighted centroids of the 2 point sets:
\begin{equation}
    \bar{\mathbf{x}} = \frac{\sum_{i=1}^{N} o_i \mathbf{x}_i}{\sum_{i=1}^{N} o_i }, 
    \qquad
    \bar{\mathbf{y}} = \frac{\sum_{i=1}^{N} o_i \mathbf{y}_i}{\sum_{i=1}^{N} o_i }.
\end{equation}

\vspace{-2mm}
\paragraph{Step 2.} Center the point clouds by subtracting away the centroid:
\begin{equation}
    \tilde{\mathbf{x}}_i = \mathbf{x}_i - \bar{\mathbf{x}}, 
    \quad 
    \tilde{\mathbf{y}}_i = \mathbf{y}_i - \bar{\mathbf{y}},
    \quad
    \forall i = 1, \dots, N.
\end{equation}

\vspace{-2mm}
\paragraph{Step 3.} Recover the rotation $\mathbf{R}$. For this, we can use the Kabsch algorithm \cite{kabsch1976svd}. First construct the following $3 \times 3$ weighted covariance matrix:
\begin{equation}
    \mathbf{H} = \sum_{i=1}^{N}{o_i \tilde{\mathbf{x}}_i \tilde{\mathbf{y}}_i^{\top}}.
\end{equation}
Considering the singular value decomposition $\mathbf{H} = \mathbf{U \Sigma V^\top}$, the desired rotation $\hat{\mathbf{R}}$ is given by:
\begin{equation}
    \mathbf{\hat{R}} = 
    \mathbf{V}
    \begin{bmatrix}
    1 & 0 & 0 \\
    0 & 1 & 0 \\
    0 & 0 & \det{\left(\mathbf{VU}^{\top}\right)}
    \end{bmatrix}
    \mathbf{U}^{\top},
\end{equation}
where $\det(\cdot)$ denotes the matrix determinant.

\vspace{-2mm}
\paragraph{Step 4.} Lastly, the translation can be computed as:
\begin{equation}
    \mathbf{\hat{t}} = \bar{\mathbf{y}} - \mathbf{\hat{R}} \bar{\mathbf{x}}.
\end{equation}

\section{Position Encodings}\label{sect:pos-emb-details}
We encode the point coordinates by generalizing the sinusoidal positional encodings in \cite{vaswani2017attention} to 3D continuous coordinates. The position encodings have the same dimension $d=256$ as the feature embeddings used in the attention layers.

For a point $\mathbf{x}=(x,y,z)$, we separately transform each coordinate to their embeddings $\mathbf{p}^{\mathbf{x}}_x, \mathbf{p}^{\mathbf{x}}_y, \mathbf{p}^{\mathbf{x}}_z \in \R^{2\floor{d/6}}$. The $x$-coordinate is transformed as:
\begin{subequations}\label{eq:posemb}
\begin{equation}
    \mathbf{p}^{\mathbf{x}}_{x}[2i] = \sin \left( \frac{x}{10000^{2i/\floor{d/3}}} \right)
\end{equation}
\vspace{-1mm}
\begin{equation}
    \mathbf{p}^{\mathbf{x}}_{x}[2i+1] = \cos \left( \frac{x}{10000^{2i/\floor{d/3}}} \right).
\end{equation}
\end{subequations}
The $y$ and $z$ coordinates are transformed in a similar manner. We then concatenate the embeddings for all three dimensions. Since the embedding dimension $d=256$ is not divisible by 6, we pad the remaining 4 elements with zeros to obtain the final embedding.

\vspace{-2mm}
\paragraph{Importance of position encodings.} 
\Cref{table:pos-enc} compares different choices of position encodings. 
The position encodings can only be removed when decoding the correspondences as a weighted sum (Eq. \ref{eq:weighted-coor}). Without position encodings, the network suffered a significant drop in performance both in terms of registration recall (RR) and accuracy (RTE and RRE), further supporting our hypothesis that our attention mechanism utilizes rigidity constraints to correct bad matches. 
We also compare with learned embeddings (using a 5-layer MLP with 32-64-128-256-256 channels), which has a slightly lower performance in most metrics. We therefore chose to use sinusoidal encodings, which also reduces the number of learnable weights. 

\begin{table}[t]
\footnotesize
\centering
\setlength\tabcolsep{3pt}
\begin{tabularx}{\linewidth}{X c | c c c | c c c}
  \hline
  & & \multicolumn{3}{c|}{\emph{3DMatch}} & \multicolumn{3}{c}{\emph{3DLoMatch}} 
  \\
  Dec. & Pos.
  & RR(\%) & RRE(\degree) & RTE(m)
  & RR(\%) & RRE(\degree) & RTE(m)
  \\
  \hline
  Wt. & None & 87.9 & 1.958 & 0.062 & 55.0 & 3.389 & 0.093
  \\
  Wt. & Learned & 89.0 & \underline{1.519} & \underline{0.047} & 61.3 & \underline{2.812} & 0.083
  \\
  Wt. & Sine & 90.9 & \textbf{1.468} & \textbf{0.046} & \underline{63.1} & \textbf{2.540} & \textbf{0.076}
  \\
  Reg. & Learned & \underline{91.6} & 1.576 & \underline{0.047} & \textbf{64.8} & 2.999 & 0.080
  \\
  \hline
  Reg. & Sine & \textbf{92.0} & 1.567 & 0.049 & \textbf{64.8} & \underline{2.827} & \underline{0.077}
  \\
  \hline
\end{tabularx}
\vspace{-2mm}
\caption{Effects of different position encodings. ``Dec.'' denotes correspondence decoding scheme, which can be either weighted coordinates using Eq.~\ref{eq:weighted-coor} (Wt.) or regression using Eq.~\ref{eq:regress-coor} (Reg.). ``Pos.'' denotes position encoding type.}
\label{table:pos-enc}
\end{table}

\begin{figure}[t]
    \centering
    \includegraphics[width=\linewidth]{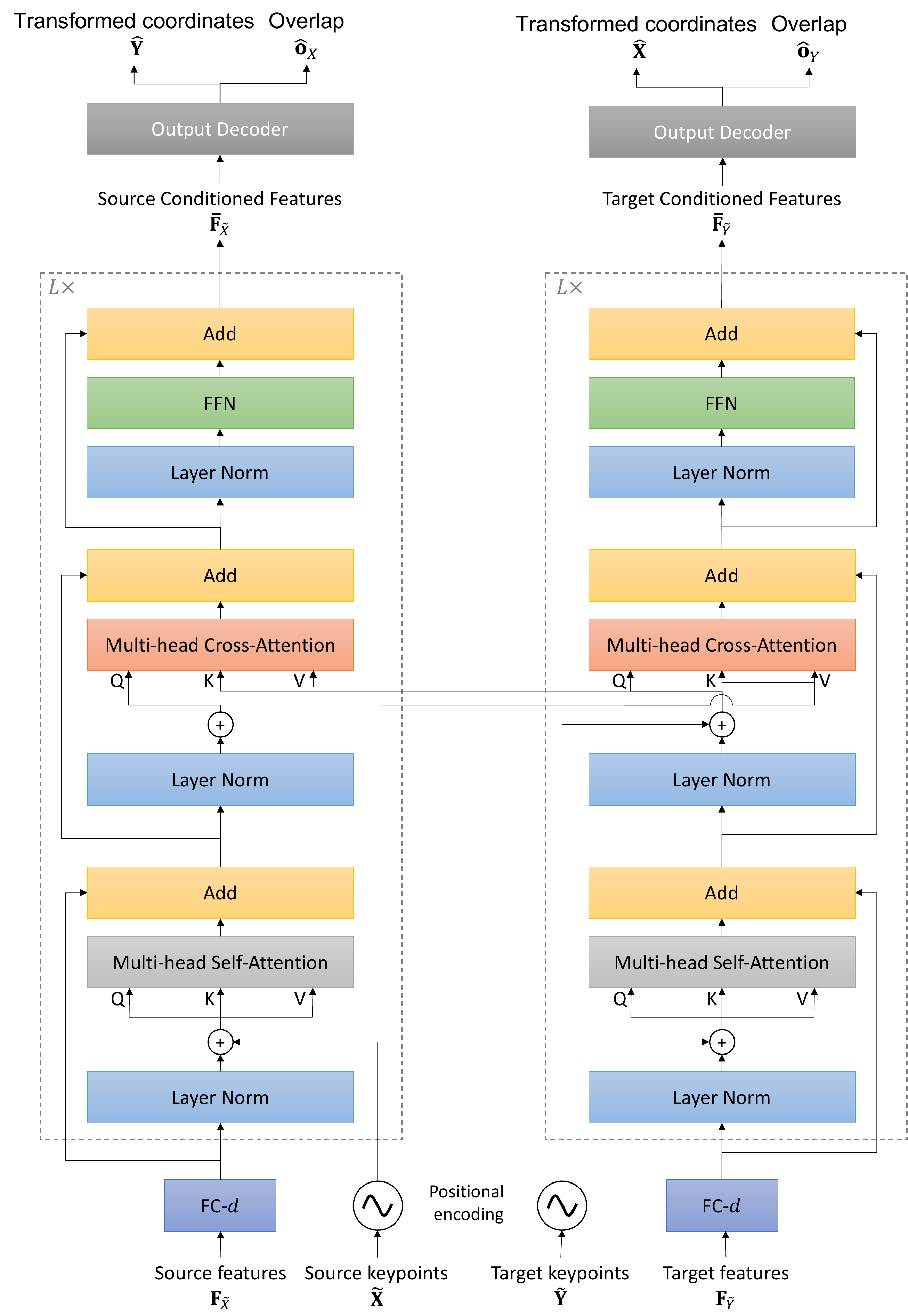}
    \caption{REGTR's transformer cross-encoder layers.}
    \label{fig:transformer-layers}
\end{figure}

\section{Network Architecture}\label{sect:network-details}
\begin{figure*}
    \centering
    \includegraphics[width=\linewidth]{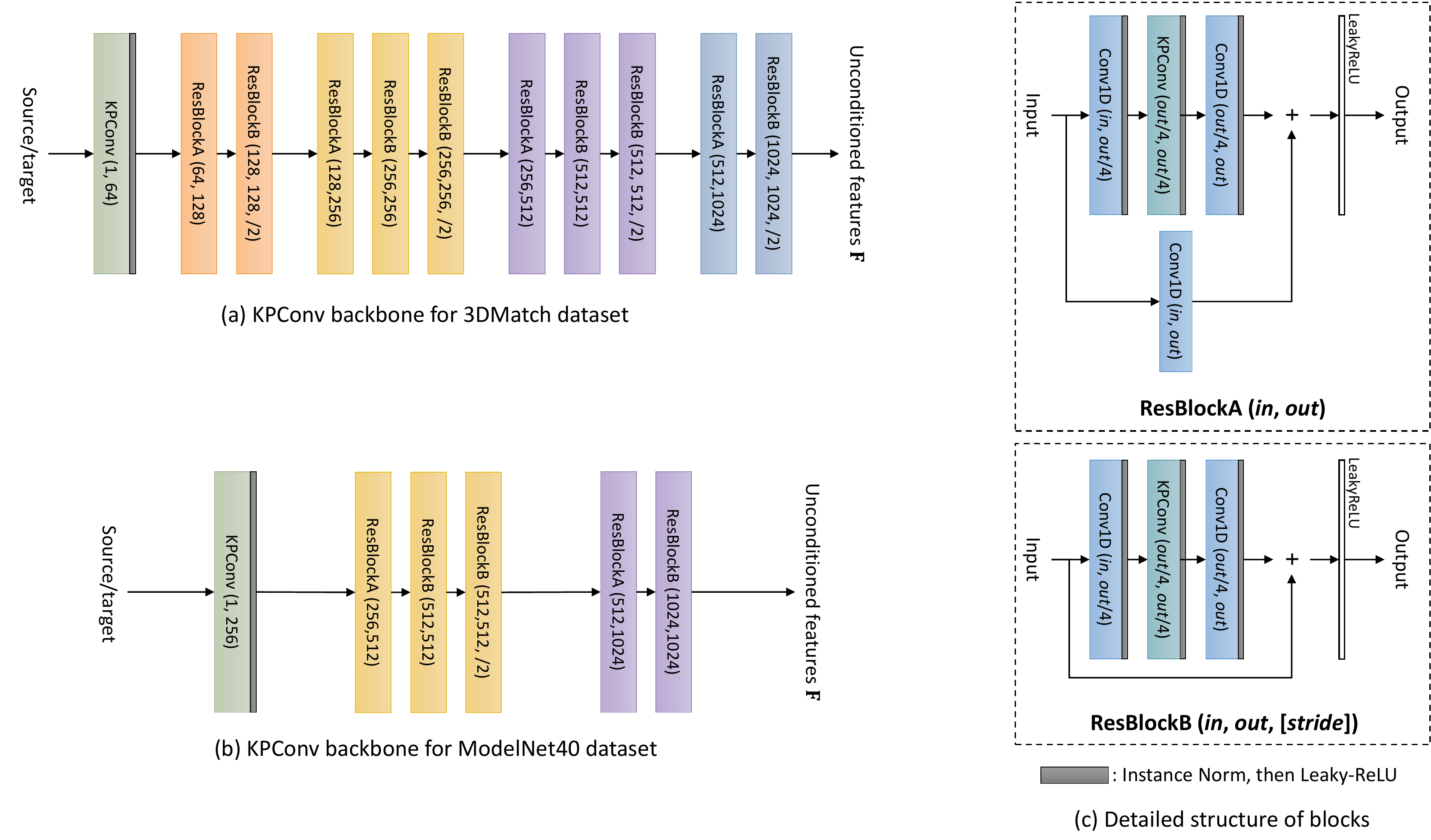}
    \caption{KPConv backbone used for (a) 3DMatch and (b) ModelNet. (c) shows the detailed structure of the residual blocks.}
    \vspace{5mm}
    \label{fig:kpconv-details}
\end{figure*}

\paragraph{KPConv backbone.}
We show the detailed network architecture of our KPConv \cite{thomas2019kpconv} backbone in \cref{fig:kpconv-details}.
We use the same KPConv backbone as Predator, but we modify it to apply instance normalization on each point cloud individually instead of over all point clouds to allow for correct behavior over batch sizes larger than one. We do not make any other changes to the backbone for the 3DMatch dataset. However, to maintain a reasonable resolution for the downsampled keypoints for ModelNet, we use a shallower backbone consisting of only a single downsampling, and the voxel size used in the first level is set to 0.03 instead of 0.06. 

\vspace{-2mm}
\paragraph{Transformer.}
\Cref{fig:transformer-layers} provides the detailed description of our transformer cross-encoder. Geometric features from the KPConv backbone are first projected to $d=256$ dimensions, and then passed through $L=6$ transformer cross-encoder layers to obtain the conditioned features $\bar{\mathbf{F}}_{\tilde{X}}, \bar{\mathbf{F}}_{\tilde{Y}}$, which can be used to predict the output correspondences and overlap scores via our output decoder.
We use the pre-LN \cite{xiong2020prenorm} configuration for the self-attention, cross-attention, and position-wise feed-forward networks (FFN). Positional encodings (Sec. \ref{sect:pos-emb-details}) are added to the queries, keys and values before every self- and cross-attention layer.

\section{Detailed Registration Results for 3DMatch}\label{sect:scannet-results-details}
We report the breakdown of the Registration Recall, Relative Rotation Error, and Relative Translation Error for each individual scene in \cref{table:3dmatch-details}. REGTR obtains the highest registration recall for three (3DMatch) and four (3DLoMatch) of the scenes, and the lowest rotation/translation errors for majority of the scenes in both settings, despite using downsampled features.

\section{Additional Qualitative Results}\label{sect:additional-qualitative}
\begin{figure*}
    \centering
    \includegraphics[width=\linewidth]{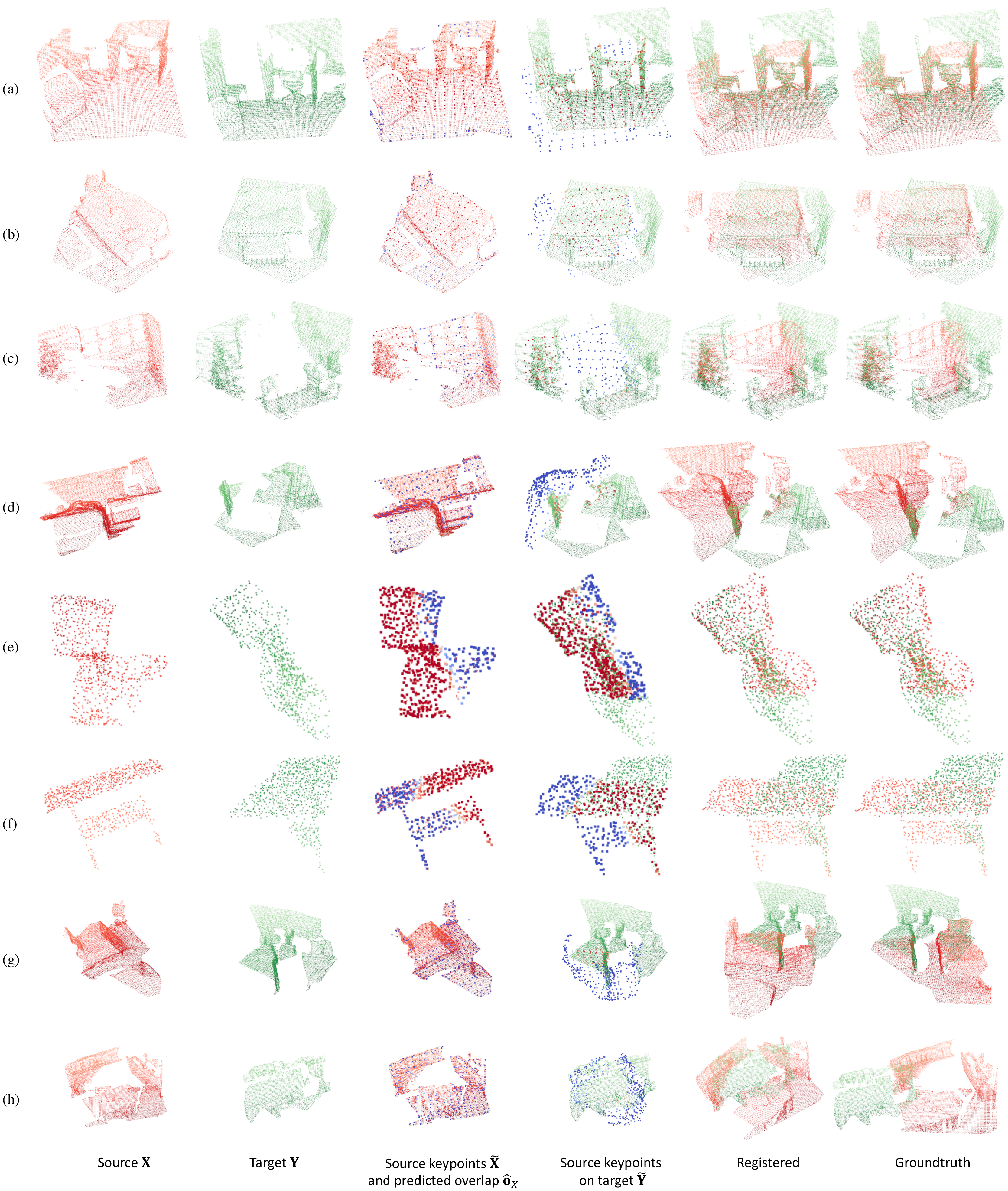}
    \caption{Additional qualitative results on (a,b) 3DMatch, (c,d) 3DLoMatch, (e) ModelNet, and (f) ModelLoNet benchmarks. Keypoints are colored by their predicted overlap scores where red indicates high overlap. The last two rows (g,h) show example failure cases on the 3DMatch dataset. Best viewed in color.}
    \label{fig:additional-qualitative}
\end{figure*}

We show additional qualitative results for both 3DMatch and ModelNet datasets in \cref{fig:additional-qualitative}. The last two rows show example failure cases. During failures, usually both overlap and correspondences are predicted wrongly.

\begin{table*}[ht]
\scriptsize
\centering
\setlength\tabcolsep{1.9pt}
\renewcommand{\arraystretch}{1.2}
\begin{tabularx}{\linewidth}{X c c c c c c c c c | c c c c c c c c c}
  \hline
  & \multicolumn{9}{c|}{3DMatch ($\geq30\%$ overlap)} & \multicolumn{9}{c}{3DLoMatch (10-30\% overlap)} \\
  & Kitchen & Home 1 & Home 2 & Hotel 1 & Hotel 2 & Hotel 3 & Study & MIT Lab & Avg.
  & Kitchen & Home 1 & Home 2 & Hotel 1 & Hotel 2 & Hotel 3 & Study & MIT Lab & Avg. \\
  
  \hline
  \multicolumn{19}{c}{\# pairs} \\
  \hline
  & 449 & 106 & 159 & 182 & 78 & 26 & 234 & 45 & 160
  & 524 & 283 & 222 & 210 & 138 & 42 & 237 & 70 & 191 \\
  
  \hline
  \multicolumn{19}{c}{Registration Recall (\%) $\uparrow$} \\
  \hline
  3DSN \cite{gojcic2019perfect}
  & 90.6 & 90.6 & 65.4 & 89.6 & 82.1 & 80.8 & 68.4 & 60.0 & 78.4
  & 51.4 & 25.9 & 44.1 & 41.1 & 30.7 & 36.6 & 14.0 & 20.3 & 33.0
  \\
  FCGF \cite{choy2019fcgf}
  & \textbf{98.0} & 94.3 & 68.6 & 96.7 & 91.0 & 84.6 & 76.1 & 71.1 & 85.1
  & 60.8 & 42.2 & 53.6 & 53.1 & 38.0 & 26.8 & 16.1 & 30.4 & 40.1
  \\
  D3Feat \cite{bai2020d3feat} 
  & 96.0 & 86.8 & 67.3 & 90.7 & 88.5 & 80.8 & 78.2 & 64.4 & 81.6
  & 49.7 & 37.2 & 47.3 & 47.8 & 36.5 & 31.7 & 15.7 & 31.9 & 37.2
  \\
  Predator-5k \cite{huang2021predator}
  & 97.6 & \underline{97.2} & 74.8 & \textbf{98.9} & \textbf{96.2} & \underline{88.5} & 85.9 & 73.3 & 89.0
  & \textbf{71.5} & 58.2 & 60.8 & \underline{77.5} & \textbf{64.2} & \underline{61.0} & 45.8 & 39.1 & 59.8
  \\
  Predator-1k \cite{huang2021predator} 
  & 97.1 & \textbf{98.1} & 74.8 & \underline{97.8} & \textbf{96.2} & \underline{88.5} & \underline{87.2} & \underline{84.4} & \underline{90.5}
  & 70.6 & \textbf{62.8} & \underline{63.1} & \textbf{80.9} & \textbf{64.2} & \underline{61.0} & \underline{50.0} & \underline{47.8} & \underline{62.5}
  \\
  Predator-NR \cite{huang2021predator} 
  & 60.8 & 74.5 & 52.2 & 80.8 & 65.4 & 57.7 & 54.3 & 55.6 & 62.7
  & 25.6 & 19.9 & 37.8 & 32.5 & 22.6 & 24.4 & 11.9 & 17.4 & 24.0
  \\
  OMNet \cite{xu2021omnet} 
  & 39.0 & 39.6 & 27.7 & 30.8 & 38.5 & 46.2 & 21.4 & 44.4 & 35.9
  & 9.0 & 6.7 & 9.0 & 3.3 & 8.8 & 14.6 & 2.5 & 13.0 & 8.4
  \\
  DGR \cite{choy2020dgr}
  & \underline{97.8} & 94.3 & 62.3 & 95.1 & 88.5 & 84.6 & 84.2 & 75.6 & 85.3
  & 60.2 & 45.0 & 52.7 & 54.5 & 48.9 & 41.5 & 38.6 & \underline{47.8} & 48.7
  \\
  PCAM \cite{cao2021pcam}
  & 96.9 & 93.4 & \textbf{78.6} & 96.7 & 83.3 & 84.6 & 81.6 & 68.9 & 85.5
  & \underline{71.1} & 56.7 & 60.8 & 70.8 & 59.9 & 41.5 & 36.4 & 42.0 & 54.9
  \\
  Ours
  & \underline{97.8} & 90.6 & \underline{75.5} & \underline{97.8} & \underline{94.9} & \textbf{100} & \textbf{88.5} & \textbf{91.1} & \textbf{92.0}
  & 66.2 & \underline{58.5} & \textbf{64.9} & 72.7 & \underline{61.3} & \textbf{70.7} & \textbf{53.0} & \textbf{71.0} & \textbf{64.8}
  \\
  \hline
  \multicolumn{19}{c}{Relative Rotation Error (\degree) $\downarrow$} \\
  \hline
  3DSN \cite{gojcic2019perfect} 
  & 1.926 & 1.843 & 2.324 & 2.041 & 1.952 & 2.908 & 2.296 & 2.301 & 2.199
  & \underline{3.020} & 3.898 & 3.427 & 3.196 & 3.217 & 3.328 & 4.325 & 3.814 & 3.528
  \\
  FCGF \cite{choy2019fcgf} 
  & \underline{1.767} & 1.849 & 2.210 & 1.867 & 1.667 & 2.417 & \underline{2.024} & 1.792 & 1.949
  & \textbf{2.904} & 3.229 & 3.277 & 2.768 & 2.801 & \textbf{2.822} & 3.372 & 4.006 & 3.147
  \\
  D3Feat \cite{bai2020d3feat}
  & 2.016 & 2.029 & 2.425 & 1.990 & 1.967 & 2.400 & 2.346 & 2.115 & 2.161
  & 3.226 & 3.492 & 3.373 & 3.330 & 3.165 & 2.972 & 3.708 & 3.619 & 3.361
  \\
  Predator-5k \cite{huang2021predator}
  & 1.861 & 1.806 & 2.473 & 2.045 & 1.600 & 2.458 & 2.067 & 1.926 & 2.029
  & 3.079 & \underline{2.637} & \textbf{3.220} & \textbf{2.694} & 2.907 & 3.390 & \underline{3.046} & 3.412 & \underline{3.048}
  \\
  Predator-1k \cite{huang2021predator} 
  & 1.902 & 1.739 & 2.306 & 1.897 & 1.817 & 2.289 & 2.278 & 2.271 & 2.062
  & 3.049 & 2.679 & 3.247 & 2.857 &\underline{2.782} & 3.340 & 3.778 & 3.538 & 3.159
  \\
  Predator-NR \cite{huang2021predator} 
  & 3.052 & 2.223 & 2.805 & 2.996 & 1.900 & 2.117 & 3.711 & 1.854 & 2.582
  & 6.445 & 4.508 & 5.272 & 5.224 & 4.889 & 6.975 & 8.457 & 5.320 & 5.886
  \\
  OMNet \cite{xu2021omnet} 
  & 4.142 & 3.166 & 3.664 & 5.450 & 3.952 & 5.518 & 4.142 & 3.296 & 4.166
  & 6.924 & 8.747 & 8.199 & 8.590 & 10.901 & 7.613 & 4.297 & \underline{3.123} & 7.299
  \\
  DGR \cite{choy2020dgr}
  & 2.181 & 1.809 & 2.474 & \underline{1.842} & 1.966 & 2.313 & 2.653 & 1.588 & 2.103
  & 4.049 & 3.967 & 4.433 & 3.666 & 4.119 & 3.742 & 4.188 & 3.469 & 3.954
  \\
  PCAM \cite{cao2021pcam}
  & 1.965 & \underline{1.644} & \underline{2.145} & 1.874 & \underline{1.434} & \textbf{1.631} & 2.250 & \underline{1.521} & \underline{1.808}
  & 3.501 & 3.518 & 3.571 & 3.649 & 3.197 & 3.278 & 4.148 & 3.368 & 3.529
  \\
  Ours
  & \textbf{1.729} & \textbf{1.347} & \textbf{1.797} & \textbf{1.639} & \textbf{1.289} & \underline{1.810} & \textbf{1.570} & \textbf{1.357} & \textbf{1.567}
  & 3.366 & \textbf{2.446} & \underline{3.244} & \underline{2.732} & \textbf{2.439} & \underline{2.919} & \textbf{3.044} & \textbf{2.428} & \textbf{2.827}
  \\
  \hline
  \multicolumn{19}{c}{Relative Translation Error (m) $\downarrow$} \\
  \hline
  3DSN \cite{gojcic2019perfect}
  & 0.059 & 0.070 & 0.079 & 0.065 & 0.074 & 0.062 & 0.093 & 0.065 & 0.071
  & 0.082 & 0.098 & 0.096 & 0.101 & \underline{0.080} & 0.089 & 0.158 & 0.120 & 0.103
  \\
  FCGF \cite{choy2019fcgf} 
  & 0.053 & 0.056 & 0.071 & \underline{0.062} & 0.061 & 0.055 & 0.082 & 0.090 & 0.066
  & 0.084 & 0.097 & \underline{0.076} & 0.101 & 0.084 & 0.077 & 0.144 & 0.140 & 0.100
  \\
  D3Feat \cite{bai2020d3feat} 
  & 0.055 & 0.065 & 0.080 & 0.064 & 0.078 & 0.049 & 0.083 & 0.064 & 0.067
  & 0.088 & 0.101 & 0.086 & 0.099 & 0.092 & 0.075 & 0.146 & 0.135 & 0.103
  \\
  Predator-5k \cite{huang2021predator} 
  & \underline{0.048} & 0.055 & 0.070 & 0.073 & 0.060 & 0.065 & \underline{0.080} & 0.063 & 0.064
  & 0.081 & \underline{0.080} & 0.084 & 0.099 & 0.096 & 0.077 & \underline{0.101} & 0.130 & \underline{0.093}
  \\
  Predator-1k \cite{huang2021predator} 
  & 0.052 & 0.062 & 0.071 & \underline{0.062} & 0.058 & 0.055 & 0.088 & 0.094 & 0.068
  & \textbf{0.077} & 0.084 & \textbf{0.074} & \textbf{0.090} & 0.093 & 0.096 & 0.126 & 0.128 & 0.096
  \\
  Predator-NR \cite{huang2021predator} 
  & 0.089 & 0.065 & 0.072 & 0.084 & 0.061 & \textbf{0.028} & 0.124 & 0.074 & 0.075
  & 0.137 & 0.106 & 0.118 & 0.158 & 0.112 & 0.152 & 0.206 & 0.193 & 0.148
  \\
  OMNet \cite{xu2021omnet} 
  & 0.103 & 0.098 & 0.097 & 0.144 & 0.099 & 0.079 & 0.124 & 0.095 & 0.105
  & 0.137 & 0.192 & 0.146 & 0.228 & 0.198 & 0.098 & 0.119 & \underline{0.094} & 0.151
  \\
  DGR \cite{choy2020dgr}
  & 0.057 & 0.064 & 0.070 & 0.068 & 0.052 & 0.062 & 0.094 & 0.072 & 0.067
  & 0.104 & 0.111 & 0.119 & 0.111 & 0.092 & 0.085 & 0.147 & 0.134 & 0.113
  \\
  PCAM \cite{cao2021pcam}
  & 0.050 & \underline{0.051} & \underline{0.060} & 0.066 & \underline{0.051} & 0.058 & 0.081 & \textbf{0.052} & \underline{0.059}
  & 0.088 & 0.108 & 0.090 & 0.112 & 0.098 & \underline{0.068} & 0.117 & 0.110 & 0.099
  \\
  Ours
  & \textbf{0.040} & \textbf{0.041} & \textbf{0.057} & \textbf{0.057} & \textbf{0.042} & \underline{0.039} & \textbf{0.054} & \underline{0.058} & \textbf{0.049}
  & \underline{0.079} & \textbf{0.064} & 0.078 & \underline{0.094} & \textbf{0.074} & \textbf{0.060} & \textbf{0.093} & \textbf{0.077} & \textbf{0.077}

  \\
  \hline
\end{tabularx}
\caption{Detailed results on the 3DMatch and 3DLoMatch datasets. Results of 3DSN, FCGF, D3Feat and Predator-5k are taken from \cite{huang2021predator}.}
\label{table:3dmatch-details}
\end{table*}

\end{document}